\documentclass[sigconf]{acmart}

\makeatletter
\newcommand*\bigcdot{\mathpalette\bigcdot@{.5}}
\newcommand*\bigcdot@[2]{\mathbin{\vcenter{\hbox{\scalebox{#2}{$\m@th#1\bullet$}}}}}
\makeatother

\definecolor{Blue}{rgb}{0, 0, 1}

\usepackage{pifont}
\renewenvironment{dinglist}[2][Blue]
{\begin{list}{\textcolor{#1}{\ding{#2}}}{}}{\end{list}}
\DeclareFontFamily{OT1}{pzc}{}%
\DeclareFontShape{OT1}{pzc}{m}{it}{<-> s * [1.200] pzcmi7t}{}%
\DeclareMathAlphabet{\mathpzc}{OT1}{pzc}{m}{it}

\usepackage{eucal}
\usepackage{mathrsfs}

\usepackage{subcaption}
\usepackage[pdftex]{floatflt}
\usepackage{graphicx}
\usepackage{multirow}
\usepackage{latexsym}

\usepackage{amsmath}
\usepackage{amssymb}

\usepackage{amsthm}
\usepackage{shadow}
\usepackage{amsbsy}
\usepackage{amsfonts}
\usepackage{enumerate}
\usepackage{moreverb}
\usepackage{array}

\usepackage{tikz}

\usepackage{color}
\usepackage{colortbl}

\newlength{\additionalvspace}%
\newlength{\colwidth}%
\newlength{\colwidthA}%
\newlength{\colwidthB}%
\newlength{\colwidthBb}%
\newlength{\colwidthC}%
\usepackage{algorithm}
\usepackage{algorithmic}
\renewcommand{\algorithmiccomment}[1]{/* #1 */}

\def\figref#1{Figure~\ref{#1}}
\def\secref#1{Section~\ref{#1}}

\def\tabref#1{Table~\ref{#1}}

\usepackage{epsfig}
\usepackage{amssymb}
\usepackage{amsmath}
\usepackage{amsfonts}

\newtheorem{property}[theorem]{Property}

\usepackage{graphicx}
\usepackage{subcaption}

\usepackage{csvsimple,booktabs}

\usepackage{etoolbox}
\newcommand{\zerodisplayskips}{%
  \setlength{\abovedisplayskip}{0pt}%
  \setlength{\belowdisplayskip}{0pt}%
  \setlength{\abovedisplayshortskip}{0pt}%
  \setlength{\belowdisplayshortskip}{0pt}}
\appto{\normalsize}{\zerodisplayskips}
\appto{\small}{\zerodisplayskips}
\appto{\footnotesize}{\zerodisplayskips}

\usepackage{hyperref}
\hypersetup{
colorlinks=true,
anchorcolor = JungleGreen,
filecolor = cyan,
menucolor=blue,
citecolor = magenta,
linkcolor=blue,
urlcolor=cyan,
bookmarks=true,
backref=true,
bookmarksopen,
bookmarksnumbered,
pagebackref=true,
hyperindex=true
}

\usepackage{dblfloatfix}
\usepackage{wasysym}
\usepackage{multicol}

\setcopyright{rightsretained}

\graphicspath{{./figs/}}

\makeatletter
\def\@copyrightspace{\relax}
\makeatother

\begin{document}

\title{On the Runtime-Efficacy Trade-off of Anomaly Detection \\
       Techniques for Real-Time Streaming Data
       }

\author{Dhruv Choudhary\ \ \ \ Arun Kejariwal\ \ \ \ Francois Orsini}
\affiliation{%
  \institution{MZ Inc.}
}

%
%

\ccsdesc[500]{Computing methodologies~Machine learning algorithms}
\ccsdesc[500]{Computer systems organization~Real-time systems}

\keywords{Stream Mining, Anomaly Detection, Time Series,
Machine Learning, Pattern Mining, Clustering}

\begin{abstract}
{\em
Ever growing volume and velocity of data coupled with decreasing attention
span of end users underscore the critical need for real-time analytics. In
this regard, anomaly detection plays a key role as an application as well as
a means to verify data fidelity. Although the subject of anomaly detection
has been researched for over 100 years in a multitude of disciplines such as,
but not limited to, astronomy, statistics, manufacturing, econometrics,
marketing, most of the existing techniques cannot be used as is on real-time
data streams.
Further, the lack of characterization of performance -- both with respect to
real-timeliness and accuracy -- on production data sets makes model selection
very challenging.

To this end, we present an in-depth analysis, geared towards real-time streaming
data, of anomaly detection techniques. Given the requirements with respect to
real-timeliness and accuracy, the analysis presented in this paper should serve
as a guide for selection of the ``best" anomaly detection technique.
To the best of our knowledge, this is the \underline{first} characterization
of anomaly detection techniques proposed in very diverse set of fields, using
production data sets corresponding to a wide set of application domains.
}
\end{abstract}

\maketitle

\vspace*{-2mm}
\section{Introduction} \label{sec:intro}

\noindent
Advances in technology -- such as, but not limited to, decreasing form factor,
network improvements and the growth of applications, such as location-based
services, virtual reality (VR) and augmented reality (AR) -- combined with
fashion to match personal styles has fueled the growth of Internet of Things
(IoT).
Example IoT devices include smart watches, smart glasses, heads-up displays
(HUDs), health and fitness trackers, health monitors, wearable scanners and
navigation devices, connected vehicles, drones et cetera.
In a recent report \cite{VNI17}, Cisco projected that, by 2021, there will be
929 M wearable devices globally, growing nearly $3\times$ from 325 M in 2016
at a CAGR of 23\%. 

The continuous and exponential increase of volume and velocity of the data
streaming from such devices limit the use of existing Big Data platforms.
To this end, recently, platforms such as Satori
\cite{Satori} have been launched to facilitate low-latency reactions on continuously
evolving data.
The other challenge in this regard pertains
to analysis of real-time streaming data. The notion of {\em real-time}, though
not new, is not well defined 
and is highly contextual \cite{kejariwal_rt}.
The following lists the various classes of latency requirements \cite{Latency}
and example applications.

\begin{figure}[!t]
\centering
\includegraphics[width=\linewidth,height=1.5in]{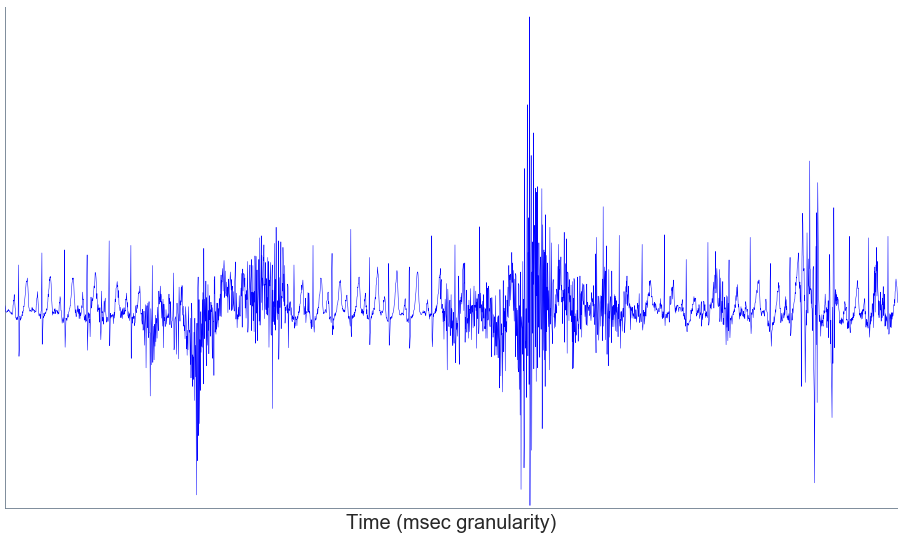}
\vspace*{-6mm}
\caption{Example ECG time series (10 msec granularity)} 
\vspace*{-5mm}
\label{fig:example}
\end{figure}

\begin{dinglist}{122}
\item {\em Nano seconds}: High Frequency Trading (HFT). 
\item {\em Micro seconds}: Data center applications,
                           cloud networks. 
\item {\em Milli seconds}: Text messaging, 
                          Publish Subscribe systems, 
                          cloud gaming. 

\item {\em 1-3 seconds}: Ad targeting.
\end{dinglist}

\begin{table*}[ht]
\centering
\resizebox{\linewidth}{!}{
\begin{tabular}{ c | l | l | c | c | c | c | c | c | c  c }
\hline
\hline
\textit{Domain} & {\textbf{Technique}} & {\textbf{Summary}} & {\textbf{P/N-P}} & {\textbf{\textless Pt,Pa \textgreater}} & {\textbf{Inc.}} & {\textbf{Robust}} & {\textbf{Recency}} & {\textbf{TG}} & {\textbf{CFAR}} \\ \hline

  \multicolumn{1}{|l|}{\cellcolor[HTML]{95A5A6}\textbf{Statistics}} & \multicolumn{9}{|l}{\cellcolor[HTML]{95A5A6}\textbf{ }} \\ \hline
  & Mu-Sigma \cite{shewhart_quality_1926, spc_shewart_quality_control, wade_review_1993, lowry_review_1995, wu_synthetic_2000, roberts_control_1959, jensen_effects_2006} & Thresholds based on mean and standard deviation & P & \textless\ding{51}, \ding{55}\textgreater & \ding{51} & & \ding{51} & 1$\mu$sec &  \\ \hline
  & Med-Mad \cite{leys_detecting_2013} & Thresholds based on median and median absolute deviation & P & \textless\ding{51}, \ding{55}\textgreater & & \ding{51} & & 1msec &  \\ \hline
  & GeneralisedESD \cite{esd_grubb_1950, grubbs_procedures_1969, tietjen_grubbs-type_1972, rosner_percentage_1983, twitter_ad_2014} & Uses Student t-distribution to calculate a max number of outliers & P & \textless\ding{51}, \ding{55}\textgreater & & \ding{51} & & 100msec &  \ding{51} \\ \hline
  & $\tau$-Estimator \cite{yohai_tau_estimator, rousseeuw_alternatives_1993, R_robustbase, qn_online} & Measure of Spread with better Gaussian efficiency than MAD & P & \textless\ding{51}, \ding{55}\textgreater & & \ding{51} &  & 1msec &  \\ \hline
  & Huber M-Estimator \cite{huber1964_mest, R_mass} & Huber's M-estimator & P & \textless\ding{51}, \ding{51}\textgreater & & \ding{51} & & 10msec &  \\ \hline
  & t-digest \cite{stream_tdigest, stream_aggr_cohen_2006, stream_sdigest} & Streaming percentile based detection & N-P & \textless\ding{51}, \ding{55}\textgreater & \ding{51} & \ding{51} & \ding{51} & 10$\mu$sec & \ding{51} \\ \hline

  & AdjBox Plots \cite{adjboxplot_hubert_2004, adjboxplot_hubert_2008} & Adjusted whiskers for box plots & N-P & \textless\ding{51}, \ding{55}\textgreater & & \ding{51} & & 10msec &  \\ \hline

  \multicolumn{1}{|l|}{\cellcolor[HTML]{95A5A6}\textbf{Time Series Analysis}} & \multicolumn{9}{|l}{\cellcolor[HTML]{95A5A6}\textbf{ }} \\ \hline
  & STL \cite{stl_cleveland_1990, twitter_ad_2014, Rsoftware, stlplus} & Seasonality Decomposition & N-P & \textless\ding{51}, \ding{55}\textgreater & & \ding{51} & & 100msec & \\ \hline
  & SARMA \cite{boxjenkins_1990, Akaike69, Akaike86, durbin_statespace, hyndman2014forecasting} & Seasonal Auto Regressive Moving Average (ARMA) & P & \textless\ding{51}, \ding{55}\textgreater & & & & 1sec & \\ \hline
  & STL-ARMA-KF \cite{ad_kalman_seminal, ad_kalman_soule_2005, ad_kalman_bera_2016, ad_kalman_vision_survey_2012, ad_kalman_survey_2010} & STL, ARMA on residuals & P & \textless\ding{51}, \ding{55}\textgreater & & & & 100msec & \\ \hline
  & STL-RobustKF \cite{ad_kalman_robust_ting_2008, ad_kalman_robust_agamennoni_2011} & ARMA with Robust Kalman & P & \textless\ding{51}, \ding{55}\textgreater & \ding{51} & \ding{51} & & 100msec & \\ \hline
  & SDAR \cite{sdar_Yamanishi_2002, sdar_Yamanishi_2004, sdar_urabe_2011} & Sequential Discounting AR  & P & \textless\ding{51}, \ding{55}\textgreater & \ding{51} & & \ding{51} & 100msec & \\ \hline
  & RobustOutlers \cite{fox_outliers_1972, muirhead_distinguishing_1986, seasonal_outliers, R_tsoutliers} & Intervention Analysis with ARMA & P & \textless\ding{51}, \ding{55}\textgreater & & \ding{51} & & 10min & \\ \hline
  & TBATS \cite{tbats_2011, holt_forecasting_2004, lucas_exponentially_1990, R_forecast, incremental_forecast_2016} & Exponential Smoothing with Fourier Terms for Seasonality & P & \textless\ding{51}, \ding{55}\textgreater & & & & 10sec & \\ \hline

  \multicolumn{1}{|l|}{\cellcolor[HTML]{95A5A6}\textbf{Pattern Mining}} & \multicolumn{9}{|l}{\cellcolor[HTML]{95A5A6}\textbf{ }} \\ \hline
  & HOTSAX \cite{pattern_hotsax_Keogh_2005, pattern_motif_Keogh_2009} & Pattern Distance based on SAX & N-P & \textless\ding{55}, \ding{51}\textgreater & \ding{51} & & & 1sec & \\ \hline
  & RRA \cite{pattern_rra_2015, pattern_grammarviz_2015} & Rare Rule Anomaly based on Grammar Induction & N-P & \textless\ding{55}, \ding{51}\textgreater & \ding{51} & & & 1sec & \\ \hline
  & DenStream \cite{denstream} & Online Density Micro-Clustering & N-P & \textless\ding{51}, \ding{51}\textgreater & \ding{51} & \ding{51} & \ding{51} & 20msec & \\ \hline
  & CluStree \cite{clustering_clustree_2011} & Hierarchical Micro-Clustering & N-P & \textless\ding{51}, \ding{51}\textgreater & \ding{51} &  &  & 100msec & \\ \hline
  & DBStream \cite{clustering_dbstream, clustering_optics_stream, inc_dbscan_1998, clustering_incr_dbscan_2013} & Incremental Shared Density Based clustering & N-P & \textless\ding{51}, \ding{51}\textgreater & \ding{51} & \ding{51} & \ding{51} & 10msec & \\ \hline

  \multicolumn{1}{|l|}{\cellcolor[HTML]{95A5A6}\textbf{Machine Learning}} & \multicolumn{9}{|l}{\cellcolor[HTML]{95A5A6}\textbf{ }} \\ \hline
  & MB$k$-means \cite{clustering_streamlda_2016, online_lda_2013} & Mini-batch clustering with $k$-means & N-P & \textless\ding{51}, \ding{51}\textgreater & & & & 10msec & \\ \hline
  & PCA \cite{Pearson1901,Hotelling33a,Hotelling33b,Jolliffe86} & Principal Components Analysis & P & \textless\ding{51}, \ding{55}\textgreater & & & & 1msec & \\ \hline
  & RobustPCA \cite{ad_rpca_candes_2011, glrm_udell_2014} & Low Rank Approximation & P & \textless\ding{51}, \ding{55}\textgreater & & \ding{51} & & 1sec & \\ \hline
  & IForest \cite{isolation_forest_2012, halfspace_forest_2011, ad_forest_guha_amazon_2016} & Isolation Forests & N-P & \textless\ding{51}, \ding{51}\textgreater & & & & 100msec & \\ \hline
  & OneSVM \cite{ad_onesvm_2001, ad_onesvm_2013} & One Label SVM & P & \textless\ding{55}, \ding{51}\textgreater & & & & 1sec & \\ \hline

  \hline
  \hline
\end{tabular}
}
\caption{Classification of anomaly detection techniques.
         {\bf P}: Parametric technique,
         {\bf N-P}: Non-parametric technique,
         {\bf Pt}: Point anomalies,
         {\bf Pa}: Pattern anomalies,
         {\bf Inc.}: Incremental technique,
         {\bf Robust}: Robustness to noise,
         {\bf Recency}: Ability to weigh observations by age,
         {\bf TG}: Time Granularity of a data stream that can use the method,
         {\bf CFAR}: Constant False Alarm Rate.
        }
\vspace*{-8mm}
\label{tab:taxonomy}
\end{table*}


\noindent
In the realm of analytics on real-time streaming data, anomaly detection plays
a key role as an application as well as a means to verify data fidelity. For
example, finding anomalies in one's vitals (see Figure \ref{fig:example} as an
illustration) can potentially help doctors to take appropriate action in a
timely fashion, thereby potentially obviating complications. Likewise, finding
anomalies in physiological signals such as electrocardiogram, electromyogram,
skin conductance and respiration of a driver can help gauge the stress level
and thereby manage non-critical in-vehicle information systems \cite{Healey05}.

Although the subject of anomaly detection has been researched for over 100 years
\cite{AD_Heron}, most of the existing techniques cannot be used as is on real-time
data streams. This stems from a multitude of reasons -- a small set of these are
mentioned below.

\begin{dinglist}{227}
\item Non-conformity between the assumptions behind a given technique and
      the underlying distribution and/or structure of data streams
\item The need for labels; in other words, many existing techniques are
      supervised in nature. Note that obtaining labels in a production
      environment is not feasible
\item Being multi-pass and non-incremental
\item Lack of support for recency
\item Lack of robustness
\item Lack of low latency computation
\item Lack of support for constant false alarm rate
\item Lack of scalability
\end{dinglist}

\noindent
Further, other characteristics of IoT devices such as, but not limited to,
small storage, small power budgets consumption et cetera limit the use of
off-the-shelf anomaly detection techniques. Last but not least, constantly
evolving nature of data streams in the wild call for support for continuous
learning.

As overviewed in \cite{AD_Heron}, anomaly detection has been researched in
a wide variety of disciplines, for example, but not limited to, operations,
computer vision, networking, marketing, and social media. Unfortunately, there
does not exist a characterization of the performance of anomaly detection
techniques -- both with respect to real-timeliness and accuracy -- on
production data sets. This in turn makes model selection very challenging.
To this end, in this paper, we present an in-depth analysis, geared towards
real-time streaming data, of a large suite of anomaly detection techniques.
In particular, the main contributions of the paper are as follows:
\begin{dinglist}{111}
\vspace{-1.5mm}
\item We present a classification of {\em over 20 (!)} anomaly detection
      techniques across seven dimensions (refer to \tabref{tab:taxonomy}).
\item As a first, using {\em over 25 (!)} real-world data sets and real
      hardware, we present a detailed evaluation of the real-timeliness
      of the anomaly detection techniques listed in \tabref{tab:taxonomy}.
      It is important to note that the evaluation was carried out in an
      unsupervised setting.
      In other words, irrespective of the availability of labels, a model
      was not trained a priori.
\item We present detailed insights into the performance -- as measured by
      {\em precision, recall} and $F_1$ score -- of the anomaly detection
      techniques listed in \tabref{tab:taxonomy}. Specifically, we also
      present a deep dive view into the behavior subject to the following:
      \begin{dinglist}{122}
      \item Trend and level shifts
      \item Change in variance
      \item Change in seasonal level
      \item Change in seasonality period
      \end{dinglist}
\item We present a map of the accuracy-runtime trade-off for the anomaly
      detection techniques.
\item Given an application domain and latency requirement, based on empirical
      evaluation, we make recommendations for the ``best" technique for
      anomaly detection.
\end{dinglist}

\noindent
Given the requirements with respect to real-timeliness and accuracy, we believe
that the analysis presented in this paper should serve as a guide for selection
of the ``best" anomaly detection technique.
To the best of our knowledge, this is the \underline{first} characterization of
anomaly detection techniques proposed in a very diverse set of fields (refer to
\tabref{tab:taxonomy}) using production data sets corresponding to a wide set of
applications.

The rest of the paper is organized as follows:
In \secref{sec:prelim}, we define the terms in the subsequent sections.
\secref{sec:back} present a brief overview of the techniques listed in Table
\ref{tab:taxonomy}.
\secref{sec:exp} details the experimental set up and \secref{sec:anal} walks
the reader through a deep dive of the analysis and insights learned from the
experiments.
%
%
Finally, in \secref{sec:conc} we conclude with directions for future work.
%
\vspace*{-2mm}

\section{Preliminaries} \label{sec:prelim}

\noindent
In this section we define the terms used in the rest of paper.

\begin{definition}
{\em Point Anomalies:} are data points which deviate so much from other data
points so as to arouse suspicions that it was generated by a different mechanism
\cite{hawkins_identification_1980}.
\label{def:1}
\vspace{-1mm}
\end{definition}

\begin{definition}
{\em Pattern Anomalies:} Continuous set of data points that are collectively
anomalous even though the individual points may or may not be point anomalies.
\label{def:2}
\vspace{-1mm}
\end{definition}

\begin{definition}
{\em Change Detection:} This corresponds to a permanent change in the structure
of a time series, e.g., change in the mean level ({\em Level Shift}), change in
the amplitude of seasonality ({\em Seasonal Level Shift}) or change in the noise
amplitude ({\em Variance Change}).
\label{def:5}
\vspace{-1mm}
\end{definition}

\begin{definition}
{\em Concept Drift:} This corresponds to the change in statistical properties,
for example, the underlying distribution, of a time series over time.
\label{def:6}
\vspace{-1mm}
\end{definition}

\noindent
Next, we define the desirable properites for anomaly detection techniques geared
towards real-time data streams.

\setcounter{theorem}{0}

\begin{property}
{\em Incremental:} A property via which a technique can analyze a new data point
without re-training a model.
\label{prop:2}
\vspace{-1mm}
\end{property}

\begin{property}
{\em Recency:} Under this, a technique assigns weights to data points which decays
with their age. In other words, recent data points play a dominant role during
model training.
\label{prop:3}
\vspace{-1mm}
\end{property}

\begin{property}
{\em Constant False Alarm Rate (CFAR):} A property under which the upper limit on
the false alarm rate (FAR) -- defined as the ratio of falsely tagged anomalies and
the total number of non-anomalous points -- is constant.
\label{prop:4}
\vspace{-1mm}
\end{property}

\vspace*{-2mm}
\section{Background} \label{sec:back}

\setcounter{secnumdepth}{3}

\noindent
As mentioned earlier in \secref{sec:intro}, the subject of anomaly detection
has been researched for over 100 years \cite{AD_Heron}. A detailed walkthrough
of prior work is beyond the scope of this paper (the reader is referred to
the books \cite{hawkins_identification_1980,barnett_outliers_1994,Rousseeuw03,
aggarwal_outlier_2013} or surveys \cite{chin_symbolic_2005,chandola_anomaly_2009,
zhang_survey_wsn_2010,gogoi_survey_2011,ndong_signal_2011,bhuyan_survey_2012,
aggarwal_outlier_temporal} written on the subject). 
%
In the section, we present a brief overview of the techniques listed in Table
\ref{tab:taxonomy}.

\vspace*{-1mm}
\subsection{Statistics}

\noindent
In this subsection we briefly overview the common statistical techniques used
for anomaly detection.

\subsubsection{Parametric Approaches}

\noindent
One of the most commonly used rule to detect anomalies -- popularly referred
to as the $\mu \pm 3 \bigcdot \sigma$ rule -- whereby, observations that lie
3 or more deviations ($\sigma$) away from the mean ($\mu$) are classified as
anomalies. The rule is based on the following two assumptions:
(a) the underlying data distribution is normal and
(b) the time series is stationary.
In practice, production time series often do not satisfy the above, which
results in false positives. Further, both $\mu$ and $\sigma$ are not robust
against the presence of anomalies. To this end, several robust estimators
have been proposed. Specifically, Huber M-estimator \cite{huber1964_mest}
is commonly used as a robust estimate of location, whereas median, $\tau$
estimator \cite{yohai_tau_estimator} and Median Absolute Deviation (MAD)
are commonly used as robust estimates of scatter.





In the presence of heavy tails in the data, $t$-distribution \cite{students_t}
is often used as an alternative to the normal distribution. The Generalized
Extreme Studentized Deviate (ESD) test \cite{rosner_percentage_1983} uses the
$t$-distribution to detect outliers in a sample, by carrying out hypothesis
tests iteratively. GESD requires an upper bound on the number of anomalies,
which helps to contain the false alarm rate (FAR).

\vspace*{-1mm}
\subsubsection{Non-parametric Approaches}

\noindent
It is routine to observe production data to exhibit, for example but not
limited to, skewed and multi-modal distribution. For finding anomalies in
such cases, several non-parametric approaches have been proposed over the
years. For instance, $t$-digest \cite{stream_tdigest} builds an empirical
cumulative density function (CDF), using adaptive bin sizes, in a streaming
fashion. Maximum bin size is determined based on the quantile of the value
$\max(1, \lfloor4N \delta q(1-q)\rfloor )$, where $q$ is the quantile and
$\delta$ is a compression factor that controls the space requirements.
%
In a similar vein, adjusted Boxplots \cite{adjboxplot_hubert_2004} have been
proposed to identify anomalies in skewed distributions. For this, it uses a
robust measure of the skew called medcouple\cite{medcouple}.

\vspace{-3mm}
\subsection{Time Series Analysis}
\vspace*{-1mm}

\noindent
Observations in a data streams exhibit autocorrelation. Thus, prior to applying
any anomaly detection technique, it is critical to weed out the autocorrelation.
Auto Regressive Moving Average (ARMA) \cite{boxjenkins_1990} models have been
commonly used for analysis of stationary time series. ARMA models are formulated
as State Space Models (SSM) \cite{durbin_statespace}, where one can employ Kalman
Filters for model estimation and inference.
Kalman filters (KF) \cite{ad_kalman_seminal} are first order Gaussian Markov
Processes that provide fast and optimal inference for SSMs. KFs assume that
the hidden and observed states are Gaussian processes. When that assumption
fails, the estimates obtained via KFs can potentially be biased. Robust Kalman
Filters ({\em RobustKF}) \cite{ad_kalman_robust_ting_2008} treat the residual
error as a statistical property of the process and down weight the impact of
anomalies on the observed and hidden states.

Sequential Discounting AutoRegressive (SDAR) filters assign more weight to
recent  observations in order to adapt to non-stationary time series or
change in dynamics of a system \cite{sdar_Yamanishi_2002}. Further, a key
feature of SDAR filters is that they update incrementally. A discount rate
is specified to guide the rate of adaptation to changing dynamics.
%

\vspace{-3mm}
\subsection{Pattern Mining}
\vspace*{-1mm}

\noindent
Time series with irregular but self-similar patterns are difficult to model
with parametric methods. Non-parametric data mining approaches that find
anomalous patterns and/or subsequences have been proposed for such time series.
SAX is a discretization technique that transforms a series from real valued
domain to a string defined over a finite alphabet $\mathcal{F}$ of size $a$
\cite{pattern_hotsax_Keogh_2005}. It divides the real number scale into equal
probabilistic bins based on the normal model and assigns a unique letter from
$\mathcal{F}$ to every bin. Before discretization, SAX z-normalizes the time
series to map it to a probabilistic scale.
It then forms words from consecutive observations that fall into a sliding
window. The time series can now be represented as a document of words. SAX
employs a dimensionality reduction technique called Piecewise Aggregate
Approximation (PAA) which chunks the time series into equal parts and computes
the average for each part. The reduced series is then discretized for further
processing.

The key advantage of SAX over other discretization heuristics \cite{haar_bu_2007,
haar_fu_2006} is that the distance between two subsequences in SAX lower bounds
the distance measure on the original series. This allows SAX to be used in
distance based anomaly detection techniques.
For example, HOTSAX \cite{pattern_hotsax_Keogh_2005} uses SAX to find the
top-$k$ discords.
%

Another method that leverages SAX is the Rare Rule Anomaly (RRA) technique
\cite{pattern_rra_2015}. RRA induces a context free grammar from the data.
The grammar induction process compresses the input sequence by learning
hierarchical grammar rules. The inability of compressing a subsequence    
is indicative of the Kolmogorov randomness of the sequence and hence, can
be treated as being an anomaly.  RRA uses the distance to the closest non-self
match subsequence as the anomaly score.

\vspace{-3mm}
\subsection{Machine Learning}
\vspace*{-1mm}

\noindent
Machine Learning approaches such as clustering, random forests, and deep learning are
very effective in modeling complex time series patterns. Having said that,
the training time is usually very high and many of these techniques are not
incremental in nature.  Thus, most of these techniques work in batch mode
where training is performed periodically.

Isolation forests \cite{isolation_forest_2012} is a tree based technique that
randomly splits the data recursively in an attempt to isolate all observations
into separate leafs. The number of splits needed to reach a data point from
the root node is called the path length. Many such random split forests are
constructed and the average path length to reach a point is used to compute
the anomaly score. Anomalous observations are closer to the root node and
hence have lower average path lengths.

One Label Support Vector Machines (SVMs) \cite{ad_onesvm_2001} are often used
to construct a non-linear decision boundary around the normal instances, thereby
isolating anomalous observation that lie away form the dense regions of the
support vectors. A key advantage of this technique is that it can be used even
with small number of data points, but are potentially slow to train.

\smallskip
\noindent {\bf Clustering}\\
\noindent
\noindent
There are two main approaches to handle the stream clustering problem:
Micro-Clustering ($MC$) and
Mini-Batching ($MB$).

\medskip
\noindent {\em Micro-Clustering} \\
\noindent
A large number of techniques follow a 2-phase Micro-clustering approach \cite{
clustering_clustream} which has both an online as well as an offline component.
The online component is a dimensionality reduction step that computes the summary
statistics for observations very close to each other (called micro-clusters).
The offline phase is a traditional clustering technique that ingests a set of
$MC$s to output the final clustering, which can be used to identify anomalous
$MC$s.

{\em DenStream} is a density-based streaming technique which uses the $MC$
paradigm \cite{denstream}. In the online phase, it creates two kinds of $MC$s:
potential micro-clusters ($p_{MC}$) and anomalous micro-clusters($o_{MC}$). Each
cluster maintains a weight $w$ which is an exponential function of the age of
the observations in the cluster. $w$ is updated periodically for all clusters
to reflect the aging of the observations. If $w$ is above a threshold
($t_{\alpha}$), it is deemed as a core micro-cluster. If $w$ is more than
$\beta t_{\alpha}$ (where $0$\textless$\beta$\textless$1$), the cluster is deemed as a $p_{MC}$,
otherwise it is deemed a $o_{MC}$. 
When a new observation arrives, the technique looks for a $p_{MC}$ that can
absorb it. If no $p_{MC}$ is found, it looks for an $o_{MC}$ that can absorb
the observation. If no $o_{MC}$ is found, a new $o_{MC}$ is instantiated.
Older clusters are periodically removed as their $w$ become smaller.
%

{\em DBStream} is an incremental clustering technique that decays $MC$s, akin
to {\em DenStream}. It also keeps track of the shared density between $MC$s.
During the offline phase, it leverages this shared density for a DBSCAN-style
clustering to identify anomalous $MC$s.

Hierarchical clustering techniques such as 
{\em Clustree} use the $MC$ paradigm to construct a hierarchical height balanced
tree of $MC$s \cite{clustering_clustree_2011}. $MC$ corresponding to an inner node
is an aggregation of the clusters of all its children.
A key advantage of these techniques is their being incremental; having said that,
the data structures can grow balloon in size. For anomaly detection, the distance
of a new observation from the closest leaf $MC$ is used as the anomaly score.

\medskip
\noindent {\em Mini-Batching} \\
\noindent
The second approach for a data stream clustering entails batch clustering of a
sample  generated from the data stream. These samples are significantly larger
than the micro-clusters and hence the name mini-batch. An example technique
of  this kind is the mini-batch $k$-means which uses cluster centroids from the
previous clustering step to reduce the convergence time significantly
\cite{minibatch_kmeans_aapl1} .

\vspace{-3mm}
\subsection{Potpourri}

\noindent
Seasonality in time series is commonly observed in production data. Filtering
of the seasonal component is critical for effective anomaly detection. In this
regard, a key challenge is how to determine the the seasonal period. For this,
a widely used approach is to detect strong peaks in the auto-correlation
function coupled with statistical tests for the strength of the seasonality.
%

Seasonal-Trend-Loess ({\em STL}) \cite{stl_cleveland_1990} is one the most
commonly used techniques for removing the seasonal and trend components. STL
uses LOESS \cite{loess} to smooth the seasonal and trend components. Presence
of anomalies can potentially induce distortions in trend which in turn can
result in false positives.
To alleviate this, {\em Robust STL} iteratively estimates the weights of the
residual -- the number of robustness iterations is an input manual parameter.
The downside of this is that the robustness iterations slow down the run time
performance. Vallis et al.\ \cite{kejariwal_stl_median} proposed the use of
piecewise medians of the trend or quantile regression -- this is significantly
faster than using the robustness iterations.
Although STL is effective when seasonality is fixed over time, moderate changes
to seasonality can be handled by choosing a lower value for the ``s.window"
parameter in the STL implementation of {\tt R}. 

Seasonal ARMA ({\em SARMA}) models handle seasonality via the use of seasonal
lag terms \cite{durbin_statespace}. A key advantage with the use of SARMA models
is the support for change in seasonality parameters over time. However, SARMA
is not robust to anomalies. This can be addressed via the use of robust filters
\cite{sarima_robust}. Note that the estimation of extra parameters increases the
relative model estimation time considerably. SARMA models can also handle multiple
seasonalities at the expense of complexity and runtime.
Akin to SARMA, {\em TBATS} is an exponential smoothing model which handles
seasonality using Fourier impulse terms \cite{tbats_2011}.

Principal Components Analysis (PCA) is another common method used to extract
seasonality. However, the use of PCA requires the data to be represented as
a matrix. One way to achieve this is to fold a time series along its seasonal
boundaries so as to build a rectangular matrix ($M$) where each row is a
complete season. Note that PCA is not robust to anomalies as it uses the
covariance matrix which is non-robust measure of scatter. To alleviate this,
Candes et al.\ proposed Robust PCA ({\em RPCA}) \cite{ad_rpca_candes_2011}, by
decomposing the matrix $M$ into a low rank component($L$) and a sparse
component($S$) by minimizing the nuclear-norm of $M$.

\vspace{-1mm}

\section{Experimental Setup} \label{sec:exp}

\noindent
In this section, we detail the data sets used for evaluation, the underlying
methodology and walk through any transformation and/or tuning needed to ensure
a fair comparative analysis.
\vspace*{-3mm}
\subsection{Data Sets} \label{sec:dsets}

\begin{figure}[!t]
\centering
\includegraphics[width=\linewidth]{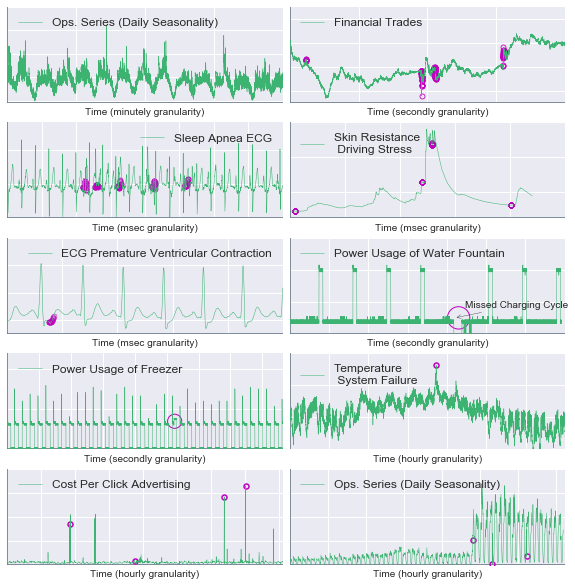}
\vspace*{-5mm}
\caption{Illustration of the characteristics of data sets from different domains}
\vspace*{-5mm}
\label{fig:datasets}
\end{figure}

\begin{table*}[ht]
\centering
\resizebox{\linewidth}{!}{
\begin{tabular}{ c | l | l | c | c | c | c | c | c | c | c | c | c | c | c c }
\hline
\hline
\textit{Domain} & {\textbf{Description}} & {\textbf{Acronym}} & {\textbf{Len}} & {\textbf{WinSize}} & {\textbf{PL}} & {\textbf{Cnt}} & {\textbf{TG}} & {\textbf{Labels}} & {\textbf{SP}}  & {\textbf{SJ}} & {\textbf{LS}} & {\textbf{VC}} & {\textbf{SLD}} & {\textbf{SLS}} \\ \hline

  \multicolumn{1}{|l|}{\cellcolor[HTML]{95A5A6}\textbf{NAB \cite{nab}}} & \multicolumn{15}{|l}{\cellcolor[HTML]{95A5A6}\textbf{ }} \\ \hline
  & NAB Advertising Click Rates & nab-ctr & 1600 & 240 & 20 & 4 & 1 hr & \ding{51} & 24 &  &  & \ding{51} & \ding{51} & \ding{51} &  \\ \hline
  & NAB Tweet Volumes & nab-twt & 15800 & 2880 & 20 & 4 & 1 hr & \ding{51} & 288 &  &  &  &  &  &  \\ \hline
  & NAB Ambient Temperature & nab-iot & 7268, 22696 & 5000 & 20 & 1 & 1 hr & \ding{51} & 24 &  & \ding{51} & \ding{51} & \ding{51} & \ding{51} \\ \hline

  \multicolumn{1}{|l|}{\cellcolor[HTML]{95A5A6}\textbf{YAD \cite{yahoo-ad}}} & \multicolumn{15}{|l}{\cellcolor[HTML]{95A5A6}\textbf{ }} \\ \hline
  & Real operations series & yahoo-a1 & 1440 & 720 & 10 & 67 & 1hr & \ding{51} & 24,168 & & \ding{51} & \ding{51} & & \ding{51} &  \\ \hline
  & Synthetic operations series & yahoo-a2 & 1440 & 720 & 10 & 100 & 1hr & \ding{51} & 100-300 & & \ding{51} & \ding{51} & & \ding{51} &  \\ \hline
  & Synthetic operations series & yahoo-a3 & 1680 & 720 & 10 & 100 & 1hr & \ding{51} & 24,168 & & \ding{51} & \ding{51} & & \ding{51} &  \\ \hline

  \multicolumn{1}{|l|}{\cellcolor[HTML]{95A5A6}\textbf{HFT \cite{quantquote}}} & \multicolumn{15}{|l}{\cellcolor[HTML]{95A5A6}\textbf{ }} \\ \hline
  & Facebook Trades Dec. 2016 & fin-fb & 334783 & 10000 & 60 & 1 & 1 sec & Manual &  &  & \ding{51} & \ding{51} & & &  \\ \hline
  & Google Trades Dec. 2016 & fin-goog & 127848 & 10000 & 60 & 1 & 1 sec & Manual &  &  & \ding{51} & \ding{51} & & &  \\ \hline
  & Netflix Trades Dec. 2016 & fin-nflx & 177018 & 10000 & 60 & 1 & 1 sec & Manual &  &  & \ding{51} & \ding{51} & & &  \\ \hline
  & SPY Trades Dec. 2016 & fin-spy & 392974 & 10000 & 60 & 1 & 1 sec & Manual &  &  & \ding{51} & \ding{51} & & &  \\ \hline

  \multicolumn{1}{|l|}{\cellcolor[HTML]{95A5A6}\textbf{Ops.}} & \multicolumn{15}{|l}{\cellcolor[HTML]{95A5A6}\textbf{ }} \\ \hline
  & Minutely Operations Data & ops-* & 120000 & 14400 & 60 & 44 & 1 min & & 1440,60 & & \ding{51} & \ding{51} & \ding{51} & \ding{51} &  \\ \hline

  \multicolumn{1}{|l|}{\cellcolor[HTML]{95A5A6}\textbf{IoT \cite{eco-ad}}} & \multicolumn{15}{|l}{\cellcolor[HTML]{95A5A6}\textbf{ }} \\ \hline
  & Power Usage of Freezer & iot-freezer01 & 432000 & 23820 & 1800 & 1 & 1 sec & & 2382 & 2370-2390 & & & \ding{51} & \ding{51} \\ \hline
  & Power Usage of Fridge & iot-fridge01 & 432000 & 81900 & 1800 & 1 & 1 sec & & 8190 & 7900-8200 & & & \ding{51} & \ding{51} \\ \hline
  & Power Usage of Dishwasher & iot-dishwasher01 & 432000 & 20000 & 1800 & 1 & 1 sec & &  &  & &  & \ding{51} & \ding{51} \\ \hline
  & Power Usage of Freezer & iot-freezer02 & 432000 & 20000 & 1800 & 1 & 1 sec & &  &  & &  & \ding{51} & \ding{51} \\ \hline
  & Power Usage of Fridge & iot-fridge02 & 432000 & 30360 & 1800 & 1 & 1 sec & & 3036 & 3030-3080 &  &  & \ding{51} & \ding{51} \\ \hline
  & Power Usage of Lamp & iot-lamp02 & 432001 & 20000 & 1800 & 1 & 1 sec & &  & 2370-2390 &  &  & \ding{51} & \ding{51} \\ \hline
  & Power Usage of Freezer & iot-freezer04 & 432000 & 41500 & 1800 & 1 & 1 sec & & 4150 & 4100-4200 &  &  & \ding{51} & \ding{51} \\ \hline
  & Power Usage of Fountain & iot-fountain05 & 432000 & 432000 & 1800 & 1 & 1 sec & & 86400 &  &  &  & \ding{51} & \ding{51} \\ \hline
  & Power Usage of Fridge & iot-fridge05 & 432000 & 47500 & 1800 & 1 & 1 sec & & 4750 & 4720-5000 &  &  & \ding{51} & \ding{51} \\ \hline

  \multicolumn{1}{|l|}{\cellcolor[HTML]{95A5A6}\textbf{Health \cite{PhysioNet}}} & \multicolumn{15}{|l}{\cellcolor[HTML]{95A5A6}\textbf{ }} \\ \hline
  & ECG Sleep Apnea \cite{apnea-ecg} & health-apnea-ecg/a02 & 15000 & 2000 & 40 & 1 & 10 msec & \ding{51} & 100 & 90-110 & \ding{51} & & & &  \\ \hline
  & ECG Seizure Epilepsy \cite{seizure} & health-szdb/sz04 & 10000 & 2000 & 40 & 1 & 5 msec & \ding{51} & 200 & 190-230 & \ding{51} & & & &  \\ \hline
  & ECG Smart Health Monitoring \cite{smart-health} & health-shareedb/02019 & 15000 & 2000 & 40 & 1 & 8 msec & \ding{51} & 110 & 95-105 & \ding{51} & & &  &  \\ \hline
  & Skin Resistance Under Driving Stress \cite{drive-stress} & health-drivedb/driver02 & 22000 & 2000 & 40 & 1 & 60 msec & \ding{51} &  &  & \ding{51} & & &  &  \\ \hline
  & Skin Resistance Under Driving Stress \cite{drive-stress} & health-drivedb/driver09/foot & 20000 & 2000 & 40 & 1 & 60 msec & \ding{51} &  &  & \ding{51} & & & &  \\ \hline
  & Respiration Under Driving Stress \cite{drive-stress} & health-drivedb/driver09/resp & 20000 & 2000 & 40 & 1 & 60 msec & \ding{51} &  &  & \ding{51} & & & &  \\ \hline
  & ECG Premature Ventricular Contraction \cite{ecg-qt}  & health-qtdb/0606 & 10000 & 2000 & 40 & 1 & 4 msec & \ding{51} & 176 & 170-180 & \ding{51} & & & &  \\ \hline

\hline
\hline
\end{tabular}
}
\caption{Anomaly detection datasets.
         {\bf Len}: Average length of the time series,
         {\bf WinSize}: Window Size,
         {\bf PL}: Pattern Length,
         {\bf Cnt}: Number of Series,
         {\bf TG}: Time Granularity,
         {\bf SP}: Seasonal Period,
         {\bf SJ}: Seasonal period Jitter,
         {\bf LS}: Level Shift,
         {\bf VC}: Variance Change,
         {\bf SLD}: Seasonal Level Drift,
         {\bf SLS}: Seasonal Level Shift
        }
\vspace*{-5mm}
\label{tab:datasets}
\end{table*}

\noindent
\tabref{tab:datasets} details the data sets used for evaluation. Note that
the data sets belong to a diverse set of domains. The diversity of the data
sets is reflected based on the following attributes:

\begin{dinglist}{122}
\item {\bf Seasonality Period} (SP): Most of these time series exhibit seasonal
      behavior and the SP increases with {\bf TG}. For instance, the minutely time
      series of operations data experience daily seasonality (period=$1440$).
      In contrast, secondly time series of {\em IoT} workloads experience
      very long seasonalities (period=$2k$ - $90k$) depending on the operation
      cycles of the appliances. {\em Health} series have coarse TG but the
      seasonal periodicity are usually small(period=$100$ - $200$).
      \vspace*{-0.5mm}
\item {\bf Seasonal Jitter}: It refers to the presence of jitter in seasonal
      periods, and is an artifact of coarse {\em TG}. {\em IoT} and {\em Health}
      time series exhibit this property.
      \vspace*{-1mm}
\item {\bf Non-stationarity}: Time series exhibit one or more types of
     nonstationarities with respect to their level (amplitude) and variance.
\end{dinglist}
\vspace*{-2mm}
\noindent
Figure \ref{fig:datasets} illustrates characteristics of these time series.
From the figure we note that most of the series are non-stationary and exhibit
one of these types of nonstationarities:


\noindent
The {\em Numenta Anomaly Benchmark} (NAB) \cite{nab} and the {\em Yahoo
Anomaly Dataset} (YAD) \cite{yahoo-ad} -- which are considered industry
standard -- provide labels. NAB itself is a collection of time series
from multiple domains like Advertising Click Through Rates ({\em nab-ctr}),
volume of tweets per hour ({\em nab-twt}), sensor data for temperature
({\em nab-iot}). The NAB series used herein have hourly granularity. YAD
is composed of three distinct data sets: {\em yahoo-a1} comprises of a set
of operations time series with hourly granularity, whereas {\em yahoo-a2}
and {\em yahoo-a3} are synthetic time series.

Anomalies detection in the context of high frequency trading (HFT) surfaces
arbitrage opportunities and hence can have potentially large impact to the
bottom line. In light of this, we included a month long time series of trades
for the tickers {\tt FB}, {\tt GOOG}, {\tt NFLX} and {\tt SPY}. The time
series, purchased from QuantQuote \cite{quantquote}, are of secondly
granularity.

We collected 44 minutely time series of operations data from our production
environment. The ECO dataset \cite{eco-ad} comprises of secondly time series
of power usage. Given the increasing emphasis on healthcare apps owing to
the use IoT devices in Healthcare domain, we included seven data sets from
Physiobank \cite{PhysioNet}.

\vspace*{-1mm}
\subsection{Methodology} \label{sec:meth}
\vspace*{-0.7mm}
\noindent
To emulate unbounded evolutionary data streams, we chose long time series and
applied anomaly detection techniques for every new data point in a streaming
fashion. Further, we limit the number of data points a technique can use to
determine whether the most recent observation is anomalous. A common way of
doing this is to define a maximum {\bf Window Size} that can be used for
training or modeling.

Window size is an important hyper-parameter which has a direct correlation with
runtime and accuracy. Longer windows require more processing time, while shorter
windows result in drop in accuracy. For a fair comparative analysis, we set an
upper limit on the values of window size for different data sets, as listed in
\tabref{tab:datasets}. The values were set based on the data set, e.g., for the
minutely operations time series (seasonal period=1440), one would need at least
$10$ periods \cite{hyndman_samplesize_2007} to capture the variance in the
seasonal patterns, giving a seasonal period of $14400$. This is the maximum
allowed valuer; techniques may choose a shorter seasonal period depending on
their requirements. For instance, {\em TBATS} needs fewer number of periods to
learn and hence, we used only $5$ periods (WinSize=$7200$) in the experiments.
Data sets such as {\em YAD} have fixed seasonal periods due to hourly {\em TG},
and hence, require much smaller window sizes to achieve maximum accuracy. {\em
IoT} data sets have the longest window sizes due to the long seasonal periods.

{\bf Pattern Length} (PL) is a hyper-parameter of all pattern mining techniques.
The value of {\em PL} is dependent on the application and the type of anomalies.
For example, {\em IoT} workloads require a PL of $\approx 30$ minutes, whereas
{\em Health} time series usually require a PL of only $40$.
A moving window equal to the {\em PL} is used to extract subsequences from the
time series. Although pattern anomalies are typically not of fixed length in
production data, most techniques require a fixed length to transform the series
into subsequences. To alleviate this, post-processing can be used to string
together multiple length anomalies \cite{pattern_rra_2015}.
%

In order to characterize the runtime performance of the technique listed in Table
\ref{tab:taxonomy}, we measured the runtime needed to process every new point in
a time series. For every new data point in a time series, each technique was run
$10$ times and the median runtime was recorded. Across a single time series, the
run-times for all the data points in the time series were averaged. Across multiple
time series in a group, geometric mean of the run-times\footnote{This approach is
an industry standard as evidenced by its use by SPEC \cite{SPECa} for over 20
years.} of the individual series is used to represent the runtime for the group.
%
%
Given that some of the data sets listed in \tabref{tab:datasets} have short time
series (e.g., {\em NAB} and {\em YAD}), we replicate these series $10$ times to
increase their length. All run-times are reported in milliseconds (msec).

\subsubsection{Metrics}

\noindent
Accuracy for labeled data sets is calculated in terms of the correctly
identified anomalies called True Positives (TP), falsely detected anomalies
called False Positives (FP) and missed anomalies called False Negatives (FN).
To measure accuracy of a single time series, we use the following three
metrics:

\vspace*{-1mm}
\begin{dinglist}{111}
\item {\bf Precision:} defined as the ratio of true positives (TP) to the total
      detected anomalies: $Pr = \frac{TP}{TP+FP}$
      \vspace*{-0.2mm}
\item {\bf Recall:} defined as the ratio of true positives (TP) to the total
      labeled anomalies: $Re = \frac{TP}{TP+FN}$
      \vspace*{-0.2mm}
\item {\bf $F_1$-score:} defined as the weighted harmonic mean of Precision and
      Recall: $F_1 = (1 + \beta^2)\frac{Pr*Re}{\beta^2 * Pr+Re}$, where $\beta$
      is a constant that weights precision vs recall based on the application
      \vspace*{-0.2mm}
\end{dinglist}

\noindent
In most applications, it is common to set $\beta$=1, giving equal weight-age to
precision and recall. But for healthcare, recall is sometimes more important
than precision and hence $\beta$=2 is often used \cite{sdar_eeg_lawhern_2013}.
This is because false negatives can be catastrophic. To calculate accuracy across
a group of time series, we report the micro-average $F_1$-score \cite{avg_f1_score},
which calculates precision and recall across the whole group. The use of this is
subject to time series in a group being similar to each other.

Most of our data sets are labeled with point anomalies. In light of this, we
propose the following methodology to compute accuracy for detected patterns
against labeled point anomalies. Let $(Y_1, Y_2, Y_3, ... Y_p, Y_{p+1})$ denote
a time series and the pattern $Y_2$-$Y_p$ be detected as a pattern anomaly. Let
$Y_{p-1}$ be a true anomaly. A naive way to compute a $TP$ is to have a pattern
anomaly end at the true anomaly. In this case, $Y_2$-$Y_p$ would be considered
a $FP$. In contrast, $TP$ can correspond to an instance where a true anomaly occurs
anywhere inside the boundary of an anomalous pattern. Pattern anomalies are very
often closely spaced due to the property of neighborhood similarity as described
by Handschin and Mayne \cite{pattern_hotsax_Keogh_2005}. Given this, it is
important to count each true anomaly only once even if multiple overlapping
pattern anomalies are detected. A post-processing pass can help weed out such
overlapping subsequences.

Given that the methodology of calculating accuracy is so different for pattern
techniques, we advise the reader to only compare accuracies of pattern techniques
with other pattern techniques.


\subsubsection{System Configuration}

\noindent
Table \ref{tab:config} details the hardware and software system configuration.
\tabref{tab:packages} details the $R$ and $Python$ packages used.
For {\em HOTSAX} and {\em RRA}, we modified the implementation so as to make
them amenable to a streaming data context. 
\begin{table}[ht]
\centering
\resizebox{\linewidth}{!}{
\begin{tabular}{c|c|c|c}
\hline
\hline
  {\cellcolor[HTML]{95A5A6}Architecture} & Intel(R) Xeon(R) & {\cellcolor[HTML]{95A5A6}Frequency} & 2.40GHz \\ \hline
  {\cellcolor[HTML]{95A5A6}Num Cores} & 24 & {\cellcolor[HTML]{95A5A6}Memory} & 128GB \\ \hline
  {\cellcolor[HTML]{95A5A6}L1d cache} & 32K & {\cellcolor[HTML]{95A5A6}L1i cache} & 32K \\ \hline
  {\cellcolor[HTML]{95A5A6}L2 cache} & 256K & {\cellcolor[HTML]{95A5A6}L3 cache} & 15360K \\ \hline
  {\cellcolor[HTML]{95A5A6}OS Version} & CentOS Linux release 7.1.1503 & {\cellcolor[HTML]{95A5A6}R, Python} & 3.2.4, 2.7 \\
\hline
\hline
\end{tabular}
}
\caption{System Configuration}
\vspace*{-8mm}
\label{tab:config}
\end{table}

\begin{table}[ht]
\centering
\resizebox{\linewidth}{!}{
\begin{tabular}{c|c|c|c}
\hline
\hline
   \multicolumn{4}{|c|}{\cellcolor[HTML]{95A5A6}\textbf{R}} \\ \hline
   \href{https://cran.r-project.org/web/packages/stream/stream.pdf}{stream}, \href{https://cran.r-project.org/web/packages/streamMOA/streamMOA.pdf}{streamMOA} & \href{https://cran.r-project.org/web/packages/MASS/MASS.pdf}{MASS} & \href{https://cran.r-project.org/web/packages/rrcov/index.html}{rrcov} & \href{https://cran.r-project.org/web/packages/jmotif/index.html}{jmotif}  \\ \hline
   \href{https://cran.r-project.org/web/packages/robustbase/index.html}{robustbase} & \href{https://cran.r-project.org/web/packages/forecast/forecast.pdf}{forecast} & \href{https://cran.r-project.org/web/packages/tsoutliers/tsoutliers.pdf}{tsoutliers}  & \href{https://cran.r-project.org/web/packages/rpca/rpca.pdf}{rpca} \\ \hline
   \multicolumn{4}{|c}{\cellcolor[HTML]{95A5A6}\textbf{Python}} \\ \hline
   \href{https://pykalman.github.io/}{pykalman} & \href{http://scikit-learn.org/stable/}{scikit-learn} & \href{https://pypi.python.org/pypi/tdigest/}{tdigest} & \href{http://www.statsmodels.org/stable/index.html}{statsmodels} \\
\hline
\hline
\end{tabular}
}
\caption{Packages and Libraries}
\vspace*{-5mm}
\label{tab:packages}
\end{table}

\vspace{-2mm}

\subsection{Hyper-parameters} \label{sec:tech}
\vspace*{-1mm}

\noindent
In the interest of reproducibility, we detail the hyper-parameters used for
the techniques listed in \tabref{tab:taxonomy}. In addition, we detail any
transformation and/or tuning needed to ensure a fair comparative analysis.

\subsubsection{Statistical techniques}

\noindent
For parametric statistical techniques such as {\em mu-sigma}, {\em med-mad}
et cetera, we set the threshold to $3\bigcdot\sigma$ or its equivalent robust
estimate of scale.

A constant false alarm rate can be set for techniques such as {\em t-digest}
and {\em GESD}.
In case of the former, we set the threshold at $99.73^{\text{th}}$  percentile
which is equivalent to $3\bigcdot\sigma$. In case of the latter, one can set an
upper limit on the number of anomalies. Based on our experiments, we set this
parameter to $0.02$ for all, but the {\em Health} data sets. The parameter was
set to $0.04$ for the the {\em Health} data sets to improve recall at the
expense of precision.

Model parameter estimates are computed at each new time step over the moving
window. As these techniques do not handle seasonality or trend, we removed
seasonality and trend using {\em STL} and evaluate these techniques on the
residual of the STL.
In light of this, i.e., statistical techniques are not applicable to the raw
time series in the presence of seasonality or trend, the accuracy of these
techniques should not be compared with ML and Pattern Mining based techniques.

\subsubsection{Time series analysis techniques}

\noindent
Parametric time series analysis techniques such as {\em TBATS}, {\em SARMA},
{\em STL-ARMA} estimate model parameters and evaluate incoming data against
the model. Retraining at every time step is often unnecessary as the temporal
dynamics do not change at every point. In practice, it is common to retrain
the model periodically and this retraining period is another hyper-parameter.
This period depends on the application, but it should not be greater than the
window size. We set the retraining period to be the same as the window size.
We include the training runtime as part of the total runtime of a techniques
so as to assess the total detection time for anomalies.

{\em STL} with default parameters assumes periodic series. To allow gradual
seasonal level drift, we set the {\em stl-periodic} parameter equal to $21$.
For {\em RobustSTL}, we use $4$ robust iterations.
{\em SDAR} is an incremental technique that requires a learning rate parameter
($r$). Based on our experiments, we set $r=0.0001$.
{\em RobustKF} is the robust kalman filter by Ting et al.\cite{ad_kalman_robust_ting_2008}
which requires parameters $\alpha$ and $\beta$ for the $Gamma$ prior on the
robustness weights. We set $\alpha = \beta = 1$.

We also evaluate techniques based on {\em Intervention Analysis} of time series
implemented in the {\em tsoutliers} package of {\em R}. These techniques are
significantly slower than most other techniques we evaluate in this work, e.g.,
for a series with $2k$ data points, it took over 5 minutes (!) for parameter
estimation. Clearly, these techniques are non-viable for real-time streaming
data.

\subsubsection{Pattern mining and machine learning techniques}

\noindent Most pattern techniques require {\em pattern length (PL)} as an input
parameter. \tabref{tab:pattern_parameters} lists the specific parameters and their
respective values for each technique. {\em HOTSAX} and {\em RRA} are robust to
presence of trend in a time series as they use symbolic approximation, but they do
require the series to be studentized. All the other pattern techniques are not robust to
presence of trend as they use the underlying real valued series directly. Thus,
all the subsequences need to be mean
adjusted (i.e., subtract the mean from all the data points) to avoid spurious
anomalies due to changing trend. Scale normalization is not carried as change in
scale is an anomaly itself.

\noindent
Keogh et al.\  proposed a noise reduction technique wherein a subsequence is
rejected if its variance lies below a threshold, $\epsilon$ \cite{pattern_hotsax_Keogh_2005}.
Our experiments show that this preprocessing step is critical, from an accuracy
standpoint, for all pattern mining techniques considerably. Therefore, we use
it by default, with $\epsilon$=$0.01$.
\vspace*{-2mm}
\vspace*{-1mm}
\begin{table}[ht]
\centering
\resizebox{\linewidth}{!}{
\begin{tabular}{l|l|l}
\hline
\hline
  {\cellcolor[HTML]{95A5A6}Technique} & {\cellcolor[HTML]{95A5A6}Parameters} & {\cellcolor[HTML]{95A5A6}Description} \\ \hline
  HOTSAX & paa-size=$4$, a-size=$5$ & PAA Size, Alphabet Size\\ \hline
  RRA & paa-size=$4$, a-size=$5$ & PAA Size, Alphabet Size\\ \hline
  DenStream & epsilon=$0.9$ lambda=$0.01$ & MC radius, Decay Constant \\ \hline
  DBStream & r=$0.5$, Cm=$5$, shared-density=True & MC radius, minimum weight for MC  \\ \hline
  ClustTree & max-height=$5$, lambda=$0.02$ & Tree Height, Decay Constant \\ \hline
  DBScan & eps=$.05$ & Threshold used to re-cluster ClusTree \\ \hline
  IForest & n-estimators=$50$, contamin.=$.05$ & Number of Trees, Number of Outliers \\ \hline
  OneSVM & nu=$0.5$, gamma=$0.1$ & support vectors, kernel coeff. \\ \hline
  MB$k$-means & n-clusters=$10$, batch-size=$200$ & Number of Clusters, Batch Size \\
\hline
\hline
\end{tabular}
}
\caption{Parameters for Pattern Mining and Machine Learning Techniques}
\vspace*{-4mm}
\label{tab:pattern_parameters}
\end{table}

\vspace*{-2mm}
{\em DenStream} is the only $MC$ technique that works without an explicit re-clustering
step. The technique can classify outlier micro-clusters ($oMC$) as anomalous, but this leads to a
higher false positive rate. Alternatively, one can take
the distance of the points to the nearest $pMC$ as the strength of the anomaly.
This makes {\em DenStream} less sensitive to the $MC$ radius $\epsilon$ as well.
{\em DenStream} and {\em DBStream} are incremental techniques. Hence, they do
not need an explicit window size parameter; having said that, they use a decay
constant $\lambda$ to discount older samples.
{\em IForest} and {\em OneSVM} are not incremental techniques and hence, need to
be retrained for each time step. {\em MBKmeans} is also not incremental, however,
it uses cluster centroids from previous run to calculate the new clusters, which allows
the clustering to converge faster.


\section{Analysis} \label{sec:anal}

\noindent
In this section we present a deep dive analysis of the techniques listed in
\tabref{tab:taxonomy}, using the data sets detailed in \tabref{tab:datasets}.

\begin{figure*}[!t]
\centering
    \begin{subfigure}[t]{0.33\textwidth}
        \raisebox{-\height}{\includegraphics[width=2.4in, height=2.5in]{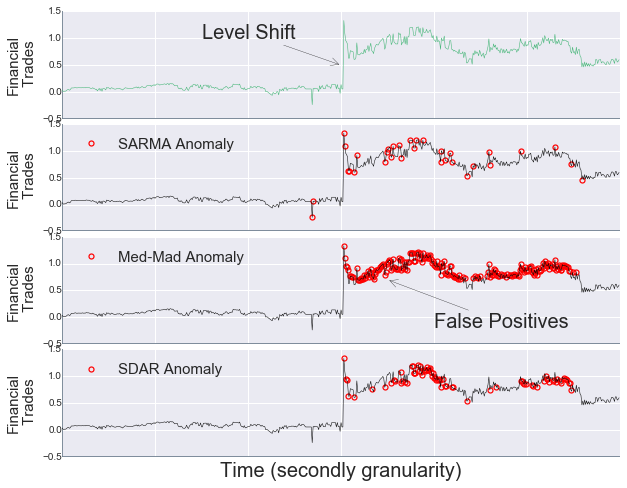}}
        \caption{An illustration of a Level Shift}
        \label{fig:levelshift}
    \end{subfigure}
    \hfill
    \hspace*{-4mm}
    \begin{subfigure}[t]{0.33\textwidth}
        \raisebox{-\height}{\includegraphics[width=2.5in, height=2.5in]{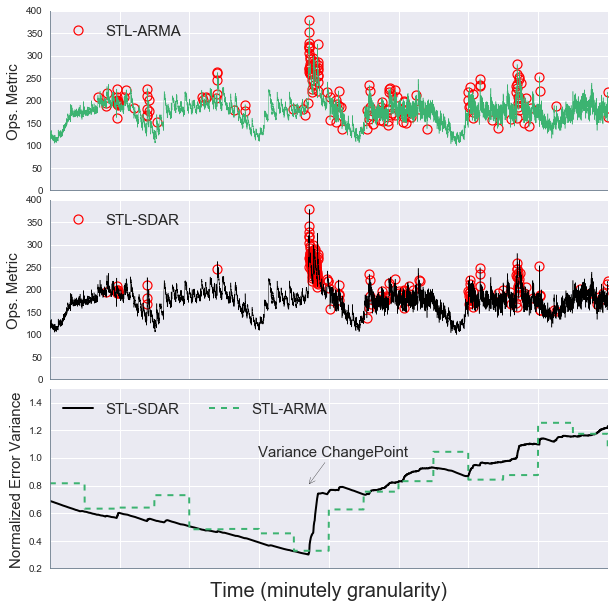}}
        \caption{An example illustrating adaptation of SDAR to change in variance}
        \label{fig:sdar_adapt}
    \end{subfigure}
    \hfill
    \hspace*{-4mm}
    \begin{subfigure}[t]{0.33\textwidth}
        \raisebox{-\height}{\includegraphics[width=2.5in, height=2.5in]{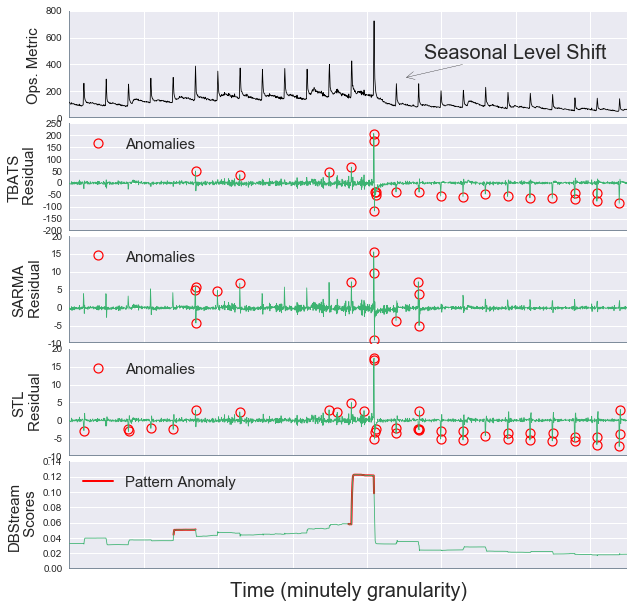}}
        \caption{Illustration of Seasonal Level Shift}
        \label{fig:tbats_sls}
    \end{subfigure}
\vspace*{-3mm}
\caption{Impact of Non-Stationarities}
\vspace*{-3mm}
\label{fig:nonstationarity}
\vspace*{-1mm}
\end{figure*}

\subsection{Handling Non-stationarity}
\vspace*{-1mm}

In this subsection we walk the reader through how the different techniques
handle the different sources of non-stationarity exhibited by the data. Most
techniques assume that the underlying process is stationary or evolving
gradually.
However, in practice, this assumption does not hold true thereby resulting
in a larger number of false positives immediately after a process change.
Though detecting the change is itself important, the false positives adversely
impact the efficacy in a material fashion.
\begin{dinglist}{112}
\item {\bf Trend and Level Shifts (LS):} Statistical techniques are not robust
      to trend or level shifts. Consequently, their performance is dependent on
      the window size, which decides how fast these techniques adapt to a new
      level.
      Time series analysis techniques based on state space models (e.g., {\em
      SARMA}, {\em TBATS}) can identify level shifts and adapt to the new level
      without adding false positives.
      \figref{fig:levelshift} illustrates a financial time series, where {\em
      SARMA} and {\em SDAR} detect the level shift as an anomaly. {\em med-mad}
      can also detect the level shift but it surfaces many false positives right
      after the level shift.
      Pattern techniques mean-adjust the patterns. Hence, in the presence of
      level shifts, they do not surface false positives as long as the pattern
      shapes do not change rapidly.
      \vspace{0.8mm}
\item {\bf Variance Change (VC):} \figref{fig:sdar_adapt} shows an operations
      series with a variance change. Iterative techniques such as {\em STL-SDAR}
      adapt faster to changing variance which allows them to limit the number
      of false positives. On the other hand, {\em STL-ARMA} and {\em SARMA} are
      periodically re-trained and oscillate around the true error variance.
      \vspace{0.8mm}
\item {\bf Seasonal Drift (SD)}: Gradually changing seasonal pattern is often
      observed in {\em iot} and {\em ops} time series. {\em SARMA} adapts to
      such a drift with default parameters. {\em STL} adapts as well if the
      $periodic$ parameter is set to false -- this ensures that seasonality
      is not averaged out across seasons.
      \vspace{0.8mm}
\item {\bf Seasonal Level Shift (SLS)}: SLS is again exhibited predominantly in
      {\em iot} and {\em ops} series as illustrated in \figref{fig:tbats_sls}.
      {\em TBATS} does not adapt to SLS or SD as it handles seasonality using
      Fourier terms and assumes that the amplitudes of seasonality do not change
      with time. {\em SARMA} handles SLS smoothly as it runs a Kalman-Filter on
      the seasonal lags and hence only detects anomalies when the shift happens.
      On the other hand, {\em STL} is not robust to SLS and may result in false
      positives as exemplified by \figref{fig:tbats_sls}.
      Pattern Techniques such as {\em DBStream} are very robust to SLS and can
      detect the pattern around the shift without any false positives.
      \vspace{0.8mm}
\item {\bf Seasonal Jitter (SJ)}: SJ is an artifact of fine-grain TG and is
      predominantly exhibited in {\em iot} (sec) and {\em health} (msec) time
      series.
      Statistical and time series analysis techniques do not model this
      non-stationary behavior. As a consequence, in such cases, only pattern
      anomaly techniques can be used.
\end{dinglist}

\vspace{-2mm}
\begin{figure*}[!t]
\centering
    \begin{subfigure}[t]{0.33\textwidth}
        \raisebox{-\height}{\includegraphics[width=2.4in, height=2.4in]{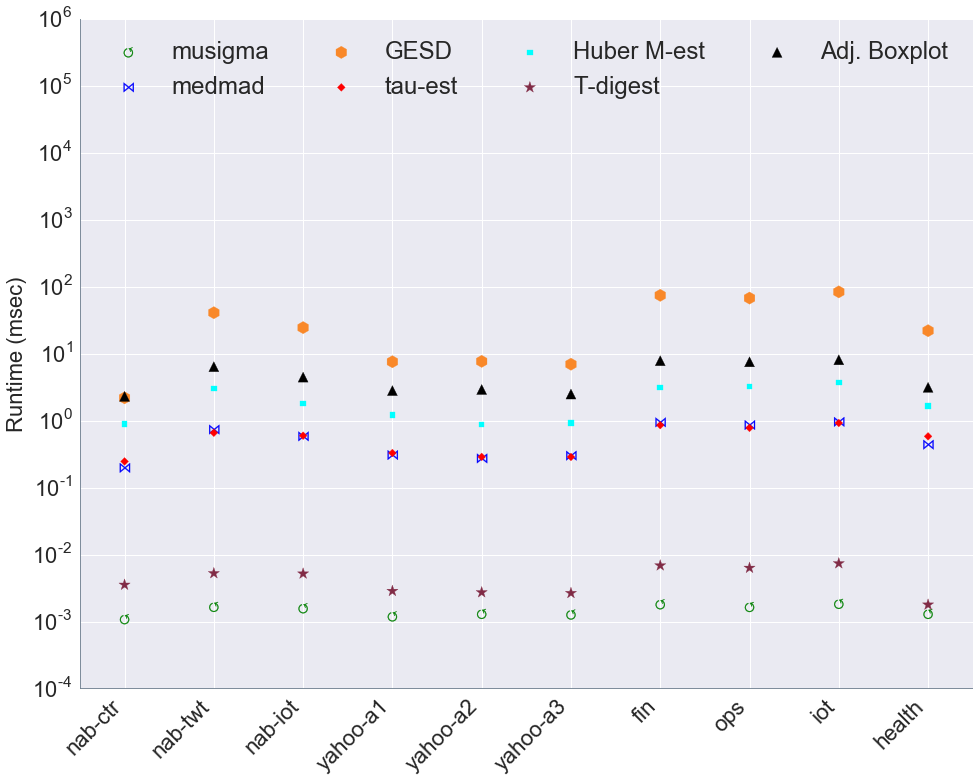}}
        \vspace*{-1mm}
        \caption{Statistical techniques}
        \vspace*{-1mm}
        \label{fig:stat_runtime}
    \end{subfigure}
    \hfill
    \hspace*{-4mm}
    \begin{subfigure}[t]{0.33\textwidth}
        \raisebox{-\height}{\includegraphics[width=2.4in, height=2.4in]{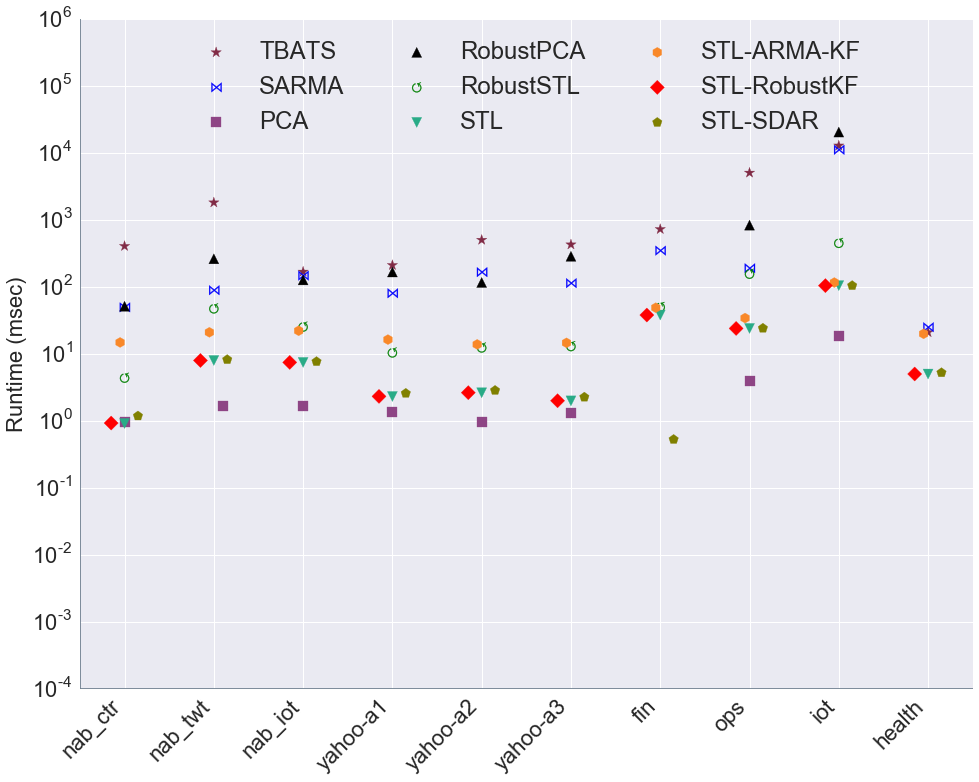}}
        \vspace*{-1mm}
        \caption{Time series analysis techniques}
        \vspace*{-1mm}
        \label{fig:ts_runtime}
    \end{subfigure}
    \hfill
    \hspace*{-4mm}
    \begin{subfigure}[t]{0.33\textwidth}
        \raisebox{-\height}{\includegraphics[width=2.4in, height=2.4in]{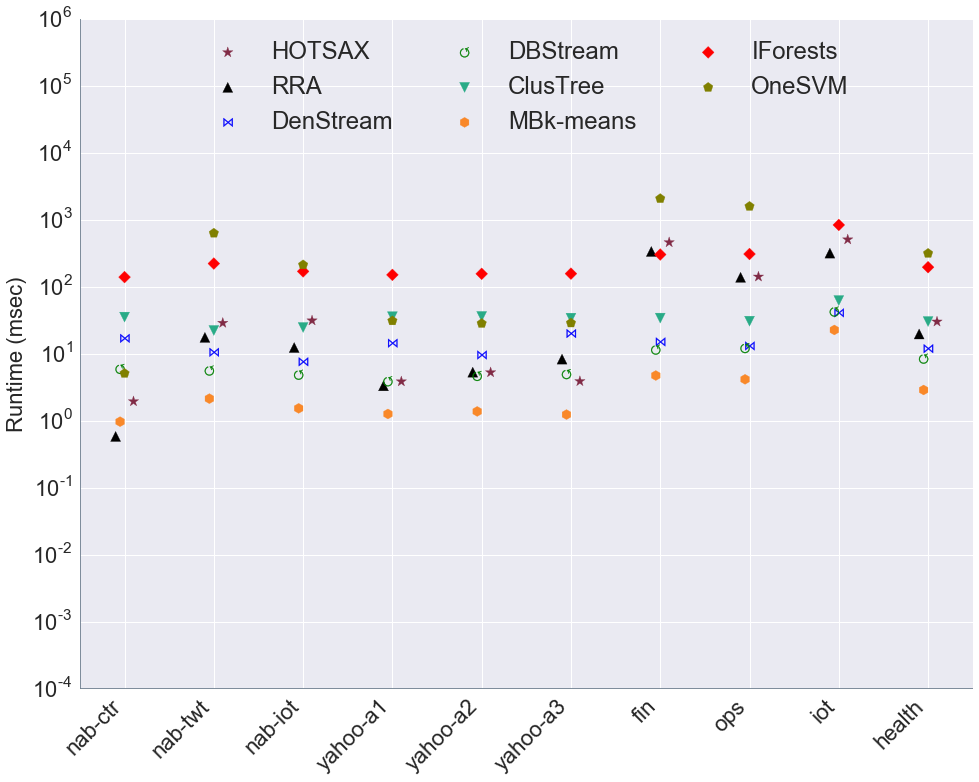}}
        \vspace*{-2mm}
        \caption{Pattern and machine learning techniques}
        \vspace*{-1mm}
        \label{fig:pattern_runtime}
    \end{subfigure}
\caption{Runtime characterization of anomaly detection techniques}
\label{fig:runtimes}
\end{figure*}

\vspace{-1mm}
\subsection{Runtime Analysis}
\vspace*{-1mm}

\noindent
In this section we present a characterization of the techniques listed in Table
\ref{tab:taxonomy} with respect to their runtimes. In the figures referenced
later in the section, the benchmarks are organized in an increasing order of
seasonal period from left to right.

\begin{dinglist}{112}
\item {\bf Statistical techniques:} From \figref{fig:stat_runtime} we note that
      {\em mu-sigma} and {\em t-digest} -- recall that these are incremental too
      -- are the fastest(<10 $\mu sec$) in this category. Robust techniques are
      at least an order of magnitude slower! This stem from the fact that these
      techniques solve an optimization problem.
      Although {\em GESD} is the slowest technique in this category, it let's one
      set an upper bound on the number of anomalies, which in turn helps control
      the false alarm rate(CFAR).
      \vspace{0.8mm}
\item {\bf Time series analysis techniques:} From \figref{fig:ts_runtime} we
      note that STL is the fastest technique (1-5 $msec$) in this category.
      Having said that, the runtime increases considerably as the seasonal
      period increases (left to right).
      {\em Robust STL} is an order of magnitude slower than {\em STL} even
      when the number of robust iterations is limited to $4$ . This can be
      ascribed to the fact that iterations with the robust weights in STL
      are significantly slower than the first one.

      \hspace*{1mm}
      {\em SARMA} and {\em TBATS} are significantly slower than most other
      techniques in this category. This is an artifact of the window length
      needed to fit these models being proportional to the seasonal period
      and thus, model parameters need to be estimated on a much larger window.
      On the other hand, a technique such as {\em STL-ARMA} applies {\em ARMA}
      on the residual of {\em STL} and therefore does not need to deal with
      seasonality, which allows for a much smaller training window. Runtimes
      for {\em TBATS, SARMA} and {\em RPCA} increase exponentially with an
      increase in seasonal period. Hence, for secondly time series, these
      methods become nonviable.

      \hspace*{1mm}
      {\em SDAR} and {\em RobustKF} are fast incremental techniques that can execute
      in $\mu secs$. However, these techniques cannot be applied to seasonal
      series as is. This limitation can be alleviated by applying STL as a
      preprocessing step. From \figref{fig:ts_runtime}, we note that {\em
      STL-SDAR} and {\em STL-RobustKF} are almost as fast as STL.
      Even though {\em STL-ARMA} trains on small training windows, note that it
      adds significant additional runtime to {\em STL}. This impact is not as
      prominent in the case of the {\em ops} and {\em fin} data sets - this is
      due to the fact that {\em STL} itself has long runtimes for these data
      sets.
      Although we note that PCA is a very fast technique, its accuracy is very
      low (this is discussed further in the next subsection). This is owing to
      the PCs being not robust to anomalies.
      themselves.
      \vspace{0.8mm}
\item {\bf Pattern and machine learning techniques:} From \figref{fig:pattern_runtime} we note that
      {\em IForest} and {\em OneSVM} have the worst runtime performance as they
      are not incremental and they need to be trained for every new data point.
      {\em MB-Kmeans} is relatively faster even though it performs clustering for
      every new point. This is because the clustering has fast convergence
      if the underlying model drift is gradual.

      \hspace*{1mm}
      Although the internal data structures {\em HOTSAX} and {\em RRA} can be
      generated incrementally, finding the farthest point using these structures
      accounts for majority of the runtime. Thus, the runtime for {\em HOTSAX,
      RRA} is not dependent on the window size; instead, it is a function of the
      variance in the data. Data sets such as {\em fin}, {\em ops} and {\em iot}
      exhibit high variance due to fine grain TG -- this impacts the runtime of
      {\em HOTSAX, RRA}. {\em med} data set on the other hand has coarse TG and
      therefore have a low variance in terms of the shapes of patterns and hence,
      HOTSAX and RRA are significantly faster for them.

      \hspace*{1mm}
      {\em DBStream} is the fastest micro-clustering technique across all data
      sets even though it does have an offline clustering component which is
      executed for every new point. This is because it maintains the shared
      density between $MC$s on-the-fly and then uses DBScan over these $MC$s
      to produce the final clustering. {\em DenStream} is slower than {\em
      DBStream} because the distance of a data point to all $pMC$'s needs to be computed
      to calculate the strength of the anomaly. Alternatively, one can tag all $oMC$s
      as anomalous which helps to reduce runtime but adversely impacts the FAR.
      From \figref{fig:pattern_runtime} we note that {\em ClusTree} is the slowest
      of all the micro-clustering techniques.

\end{dinglist}

\begin{figure}[!h]
\centering
\includegraphics[width=\linewidth, height=2.5in]{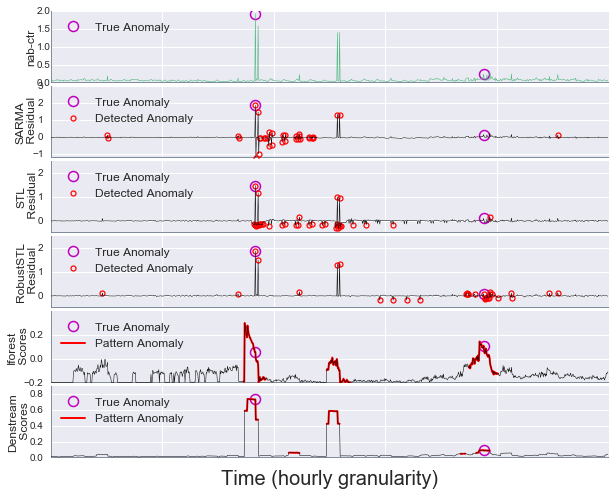}
\vspace*{-7mm}
\caption{Illustration of the impact of lack of robustness}
\vspace*{-7mm}
\label{fig:robustness}
\end{figure}

\vspace{-1mm}
\subsection{Accuracy-Speed Tradeoff}

\begin{figure*}[!t]
\centering
    \begin{subfigure}[t]{0.33\textwidth}
        \raisebox{-\height}{\includegraphics[width=\textwidth]{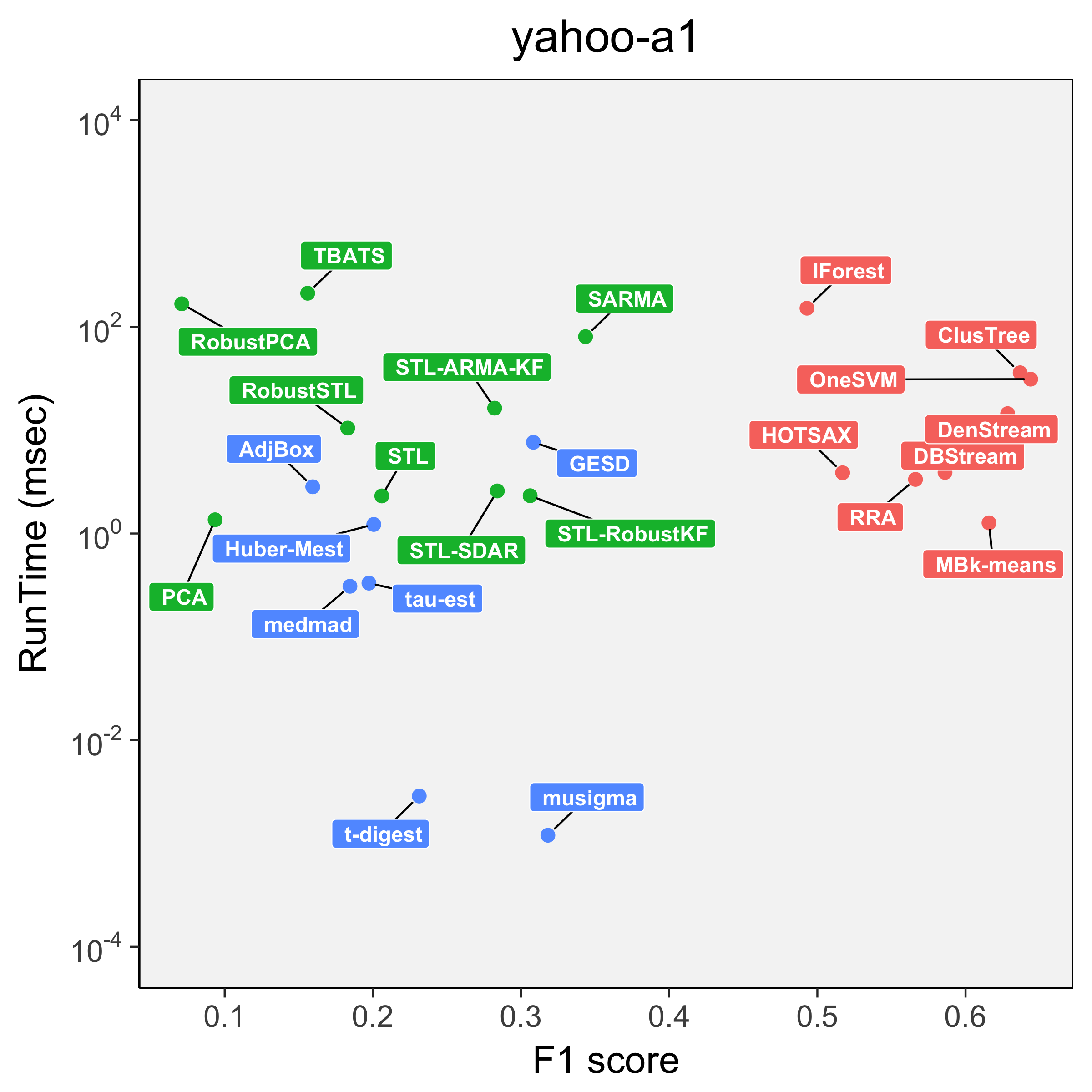}}
        \vspace*{-0.5mm}
    \end{subfigure}
    \hfill
    \hspace*{-4mm}
    \begin{subfigure}[t]{0.33\textwidth}
        \raisebox{-\height}{\includegraphics[width=\textwidth]{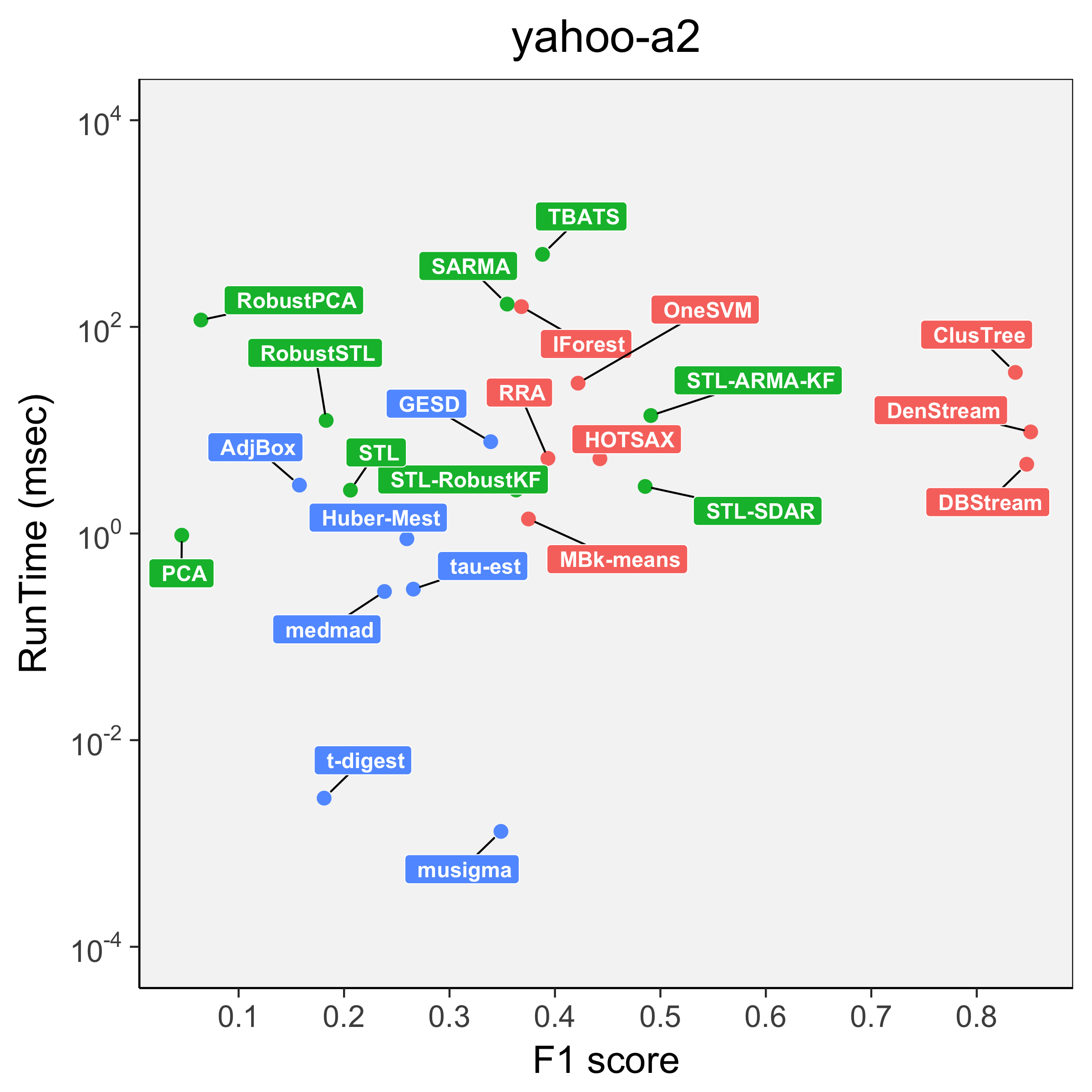}}
        \vspace*{-0.5mm}
    \end{subfigure}
    \hfill
    \hspace*{-4mm}
    \begin{subfigure}[t]{0.33\textwidth}
        \raisebox{-\height}{\includegraphics[width=\textwidth]{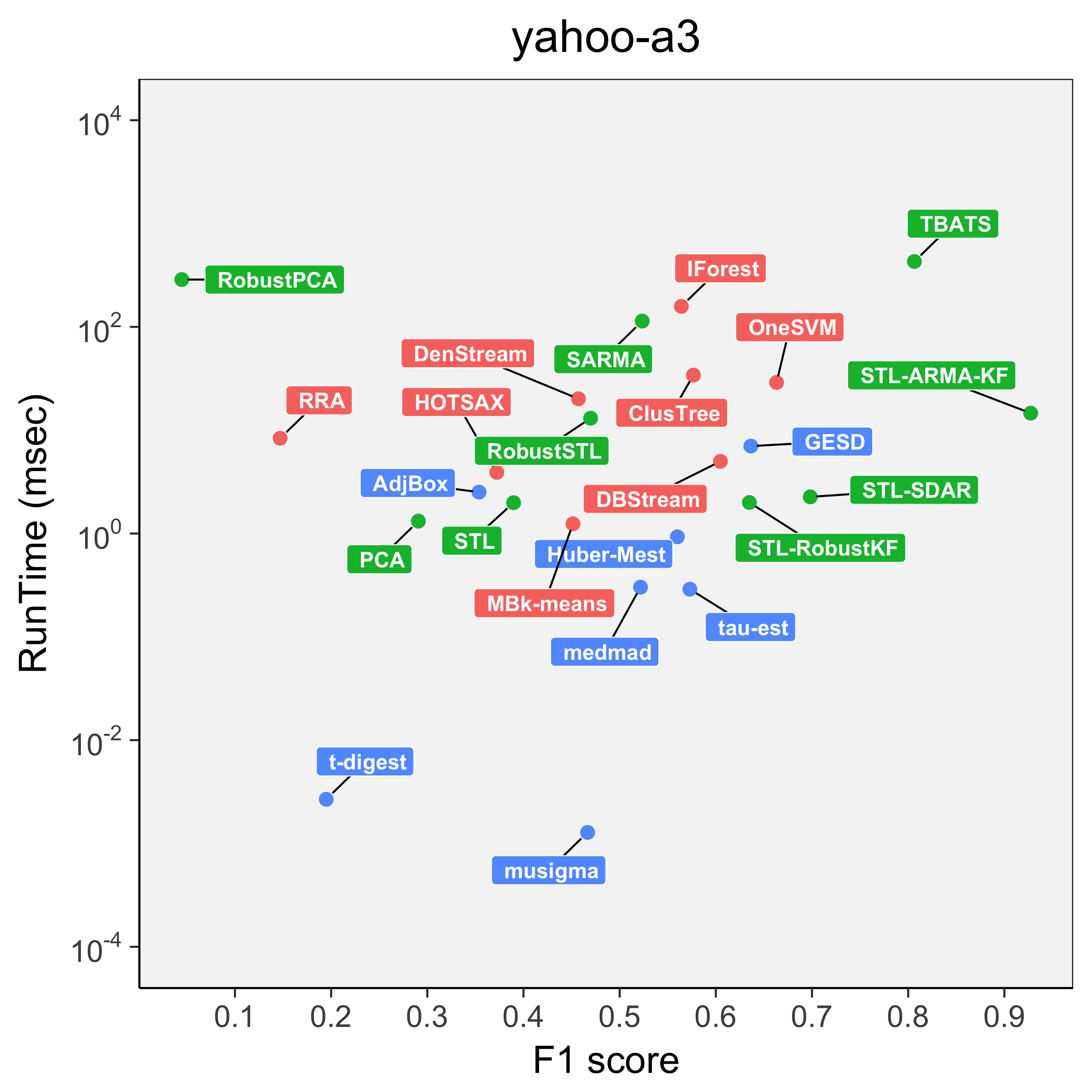}}
        \vspace*{-0.5mm}
    \end{subfigure}
    \begin{subfigure}[t]{0.33\textwidth}
        \raisebox{-\height}{\includegraphics[width=\textwidth]{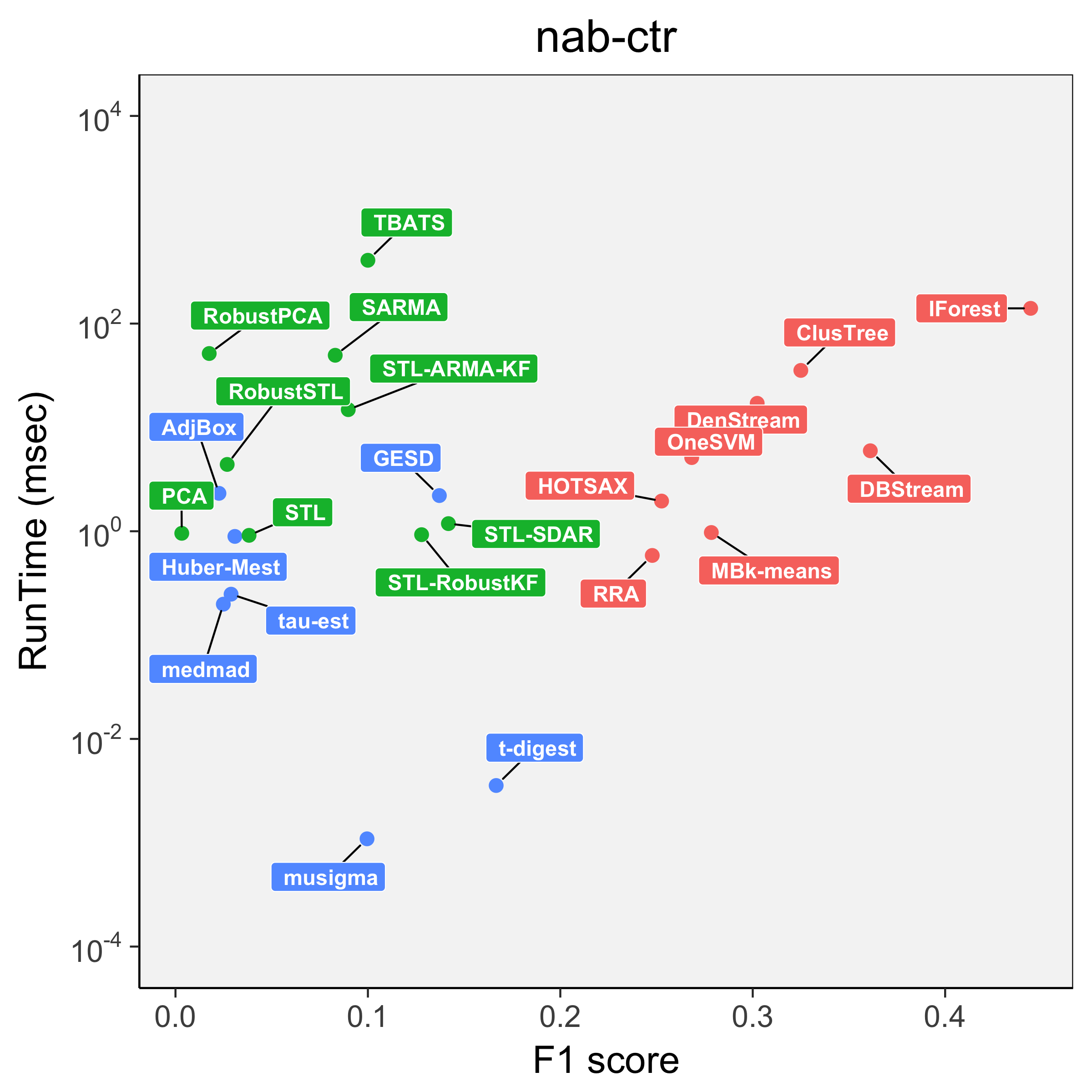}}
        \vspace*{-0.5mm}
    \end{subfigure}
    \hfill
    \hspace*{-4mm}
    \begin{subfigure}[t]{0.33\textwidth}
        \raisebox{-\height}{\includegraphics[width=\textwidth]{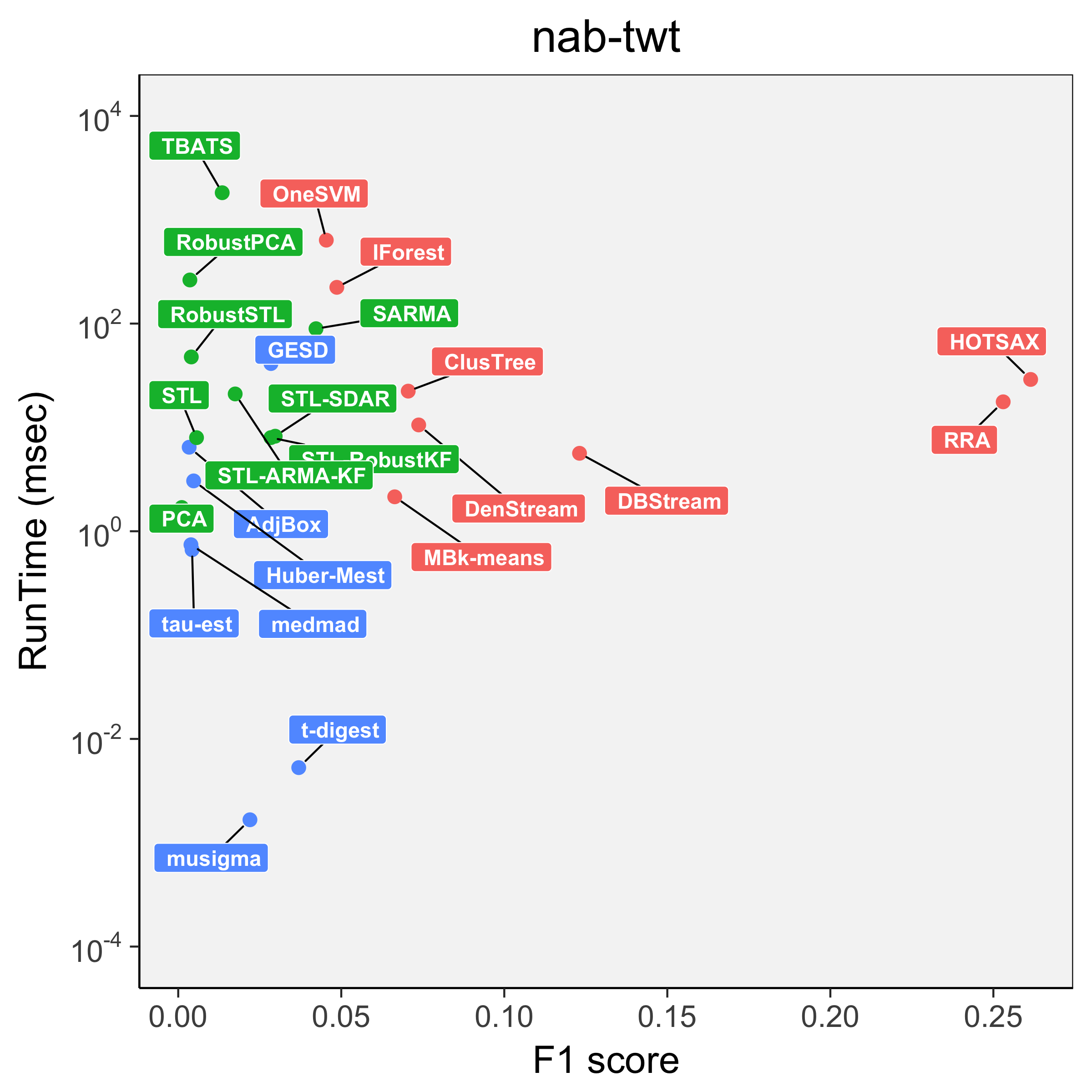}}
        \vspace*{-0.5mm}
    \end{subfigure}
    \hfill
    \hspace*{-4mm}
    \begin{subfigure}[t]{0.33\textwidth}
        \raisebox{-\height}{\includegraphics[width=\textwidth]{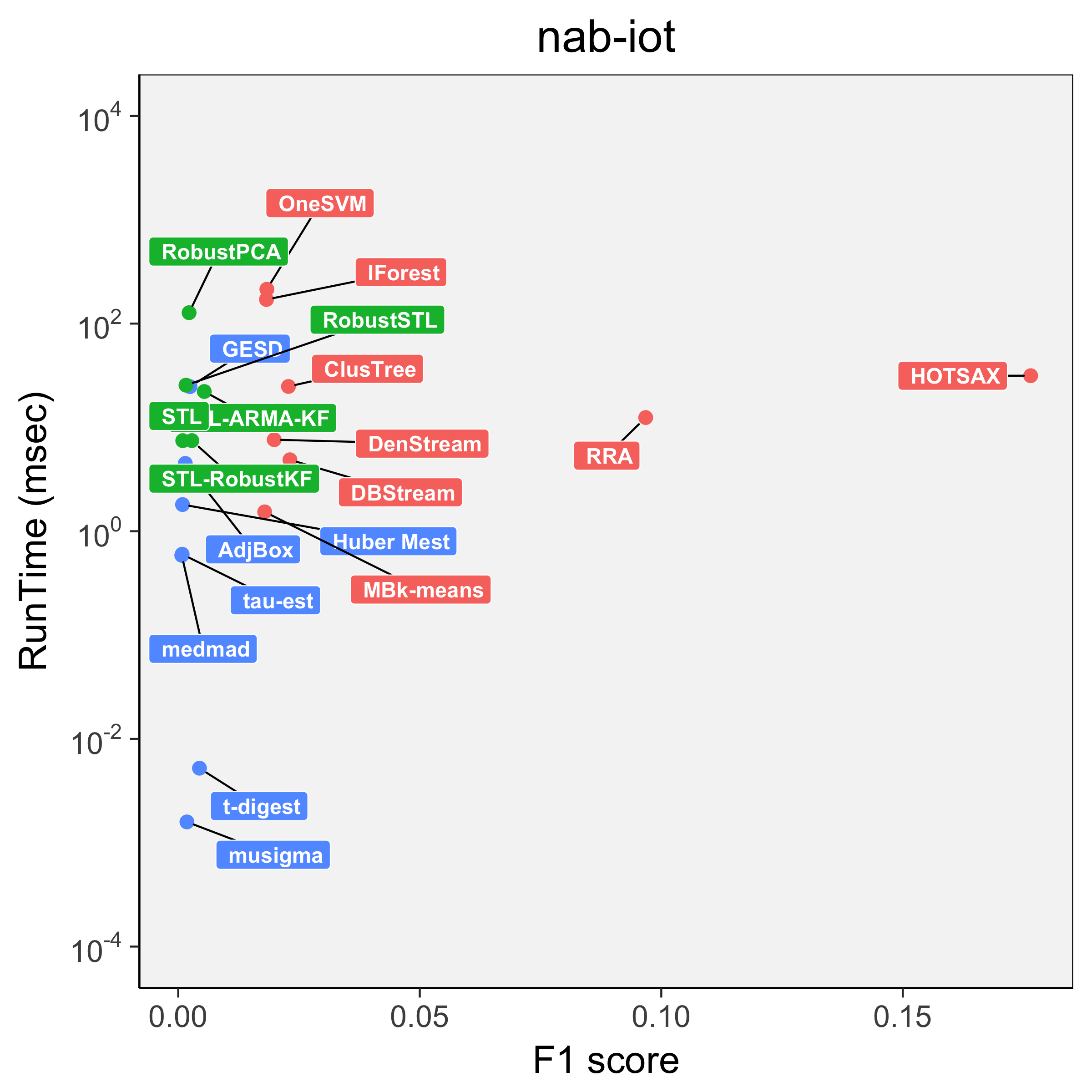}}
        \vspace*{-0.5mm}
    \end{subfigure}
    \begin{subfigure}[t]{0.33\textwidth}
        \raisebox{-\height}{\includegraphics[width=\textwidth]{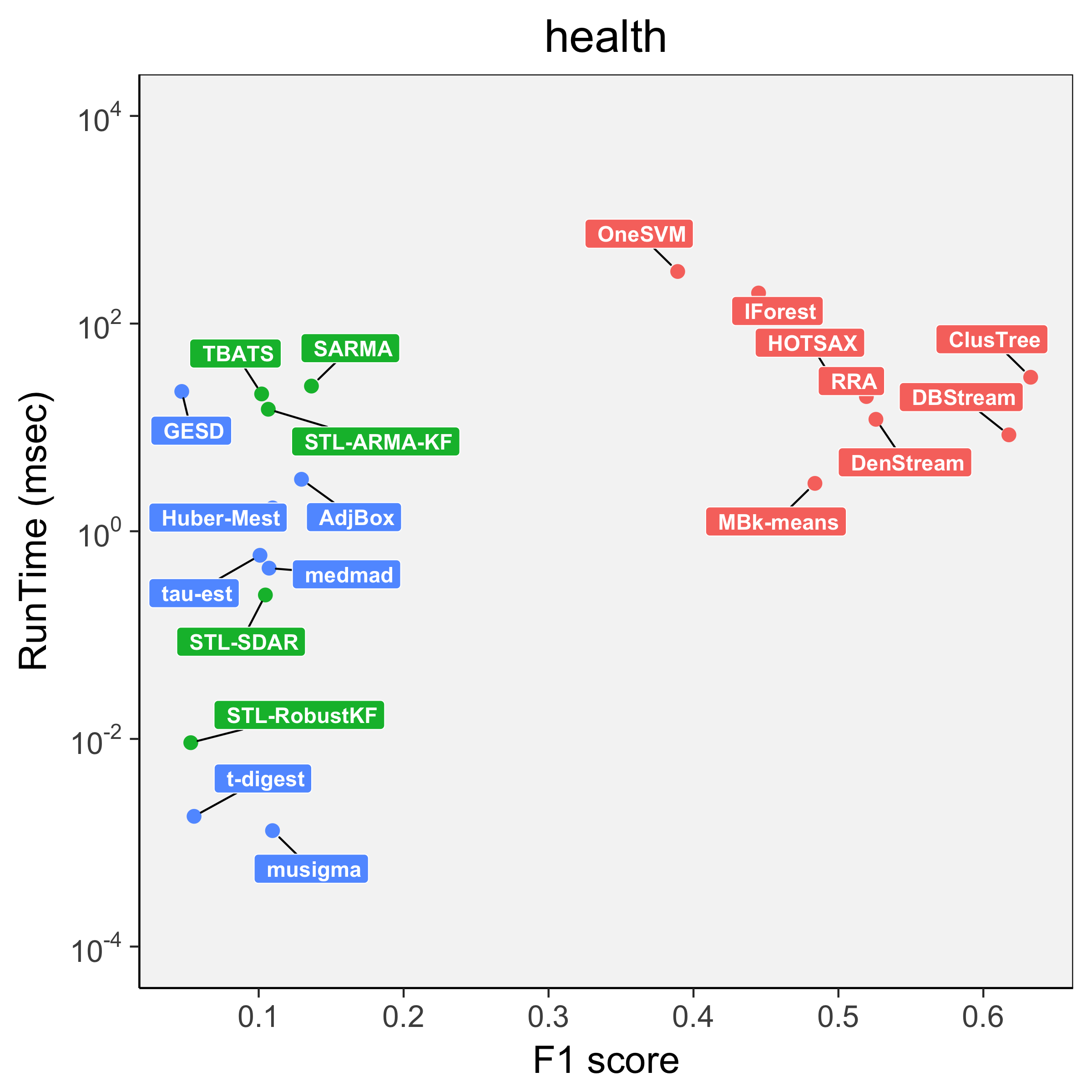}}
        \vspace*{-0.5mm}
    \end{subfigure}
    \begin{subfigure}[t]{0.33\textwidth}
        \raisebox{-\height}{\includegraphics[width=\textwidth]{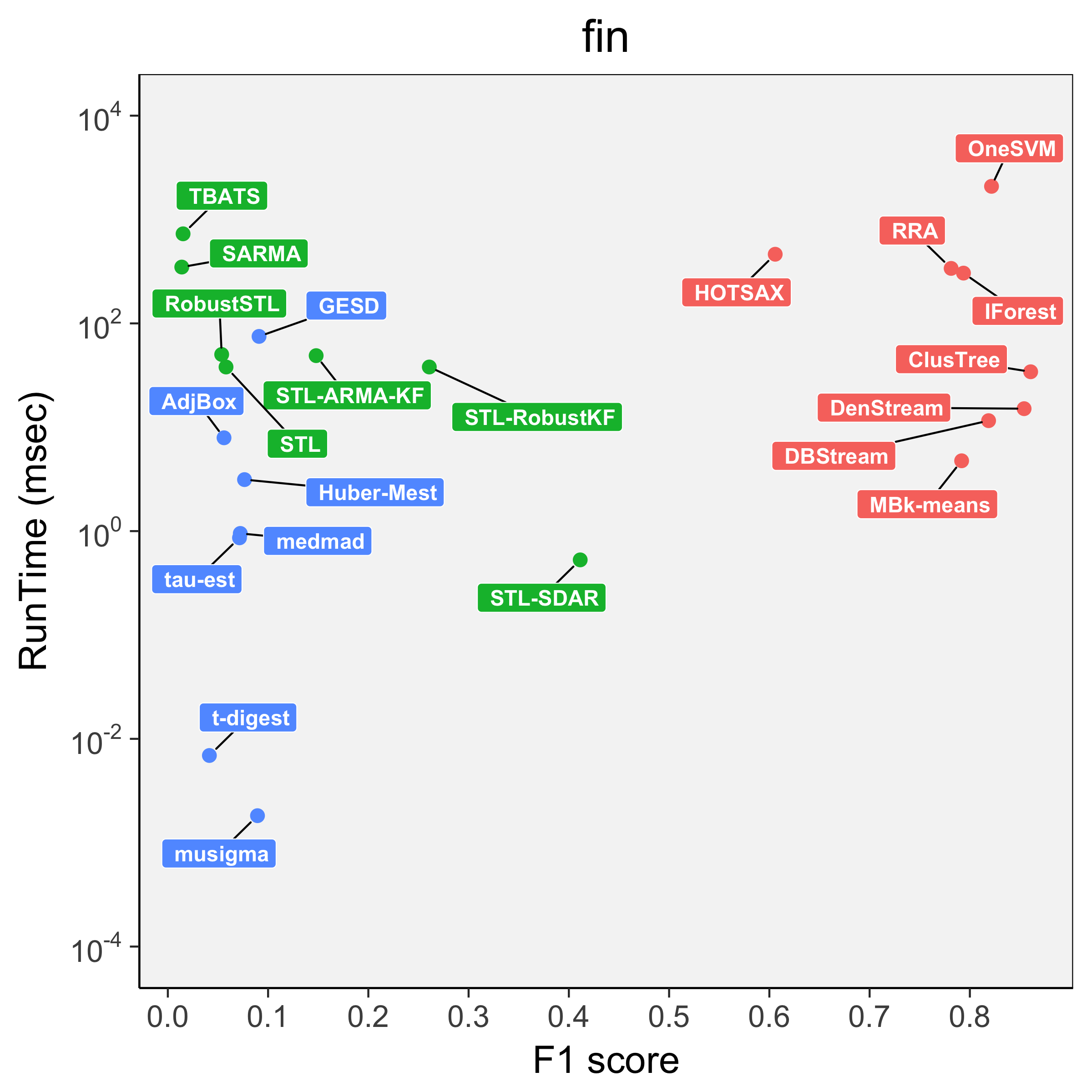}}
        \vspace*{-0.5mm}
    \end{subfigure}
\vspace*{-2mm}
\caption{Landscape of accuracy-speed tradeoff.
         Note that Y-axis is on log scale.
         For ease of visual analysis, the range for X-axis is custom for each dataset.
         Techniques are grouped according to scale -  Blue: Statistical, Green: Time Series, Red: Pattern
        }
\vspace*{-5mm}
\label{fig:accspeed}
\end{figure*}

\noindent
Figure \ref{fig:accspeed} charts the landscape of accuracy-speed tradeoff for
the techniques listed in \tabref{tab:taxonomy} across all data sets tabulated in
\tabref{tab:datasets}. \tabref{tab:accuracytable} details the {\em Precision},
{\em Recall} along with the {\em $F_1$-score}. In this rest of this subsection, we
present an in-depth analysis of the trade-off from multiple standpoints.


\begin{table*}[ht]
\centering
\resizebox{\linewidth}{!}{
\begin{tabular}{l||c|c|c||c|c|c||c|c|c||c|c|c||c|c|c||c|c|c||c|c|c||c|c|c}
\hline
\hline

\textit{Datasets} & \multicolumn{3}{|c|}{\cellcolor[HTML]{95A5A6}\textbf{MuSigma}} & \multicolumn{3}{|c|}{\cellcolor[HTML]{95A5A6}\textbf{Med-MAD}} & \multicolumn{3}{|c|}{\cellcolor[HTML]{95A5A6}\textbf{tau-est}} & \multicolumn{3}{|c|}{\cellcolor[HTML]{95A5A6}\textbf{Huber Mest}} & \multicolumn{3}{|c|}{\cellcolor[HTML]{95A5A6}\textbf{AdjBox}} & \multicolumn{3}{|c|}{\cellcolor[HTML]{95A5A6}\textbf{GESD}} & \multicolumn{3}{|c|}{\cellcolor[HTML]{95A5A6}\textbf{t-digest}} & \multicolumn{3}{|c|}{\cellcolor[HTML]{95A5A6}\textbf{SARMA}} \\ \hline
\hline

         & {\textbf{Pr}} & {\textbf{Re}} & {\textbf{$F_1$}} & {\textbf{Pr}} & {\textbf{Re}} & {\textbf{$F_1$}} & {\textbf{Pr}} & {\textbf{Re}} & {\textbf{$F_1$}} & {\textbf{Pr}} & {\textbf{Re}} & {\textbf{$F_1$}} & {\textbf{Pr}} & {\textbf{Re}} & {\textbf{$F_1$}} & {\textbf{Pr}} & {\textbf{Re}} & {\textbf{$F_1$}} & {\textbf{Pr}} & {\textbf{Re}} & {\textbf{$F_1$}} & {\textbf{Pr}} & {\textbf{Re}} & {\textbf{$F_1$}} \\ \hline

yahoo-a1 & 0.268 & 0.392 & 0.318 & 0.108 & 0.623 & 0.185 & 0.118 & 0.61  & 0.197 & 0.121 & 0.592 & 0.201 & 0.092 & 0.579 & 0.159 & 0.303 & 0.314 & 0.308 & 0.221 & 0.242 & 0.231 & 0.284 & 0.434 & 0.344 \\ \hline
yahoo-a2 & 0.234 & 0.687 & 0.349 & 0.139 & 0.83  & 0.238 & 0.158 & 0.83  & 0.266 & 0.154 & 0.83  & 0.26  & 0.087 & 0.83  & 0.158 & 0.215 & 0.803 & 0.339 & 0.107 & 0.588 & 0.181 & 0.232 & 0.749 & 0.355 \\ \hline
yahoo-a3 & 0.467 & 0.745 & 0.467 & 0.353 & 0.999 & 0.522 & 0.402 & 0.999 & 0.573 & 0.389 & 0.999 & 0.56  & 0.215 & 0.999 & 0.354 & 0.47  & 0.986 & 0.636 & 0.111 & 0.787 & 0.195 & 0.355 & 0.993 & 0.523 \\ \hline
nab-ctr  & 0.053 & 0.786 & 0.1   & 0.013 & 1     & 0.025 & 0.015 & 1     & 0.029 & 0.016 & 1     & 0.031 & 0.011 & 1     & 0.023 & 0.075 & 0.857 & 0.137 & 0.093 & 0.786 & 0.167 & 0.044 & 0.786 & 0.083 \\ \hline
nab-twt  & 0.011 & 0.743 & 0.022 & 0.002 & 0.914 & 0.004 & 0.002 & 0.914 & 0.004 & 0.002 & 0.914 & 0.005 & 0.002 & 0.886 & 0.003 & 0.015 & 0.714 & 0.028 & 0.019 & 0.743 & 0.037 & 0.022 & 0.75  & 0.042 \\ \hline
nab-iot  & 0.001 & 0.167 & 0.002 & 0     & 0.167 & 0.001 & 0     & 0.167 & 0.001 & 0     & 0.167 & 0.001 & 0.001 & 0.333 & 0.001 & 0.001 & 0.167 & 0.002 & 0.002 & 0.333 & 0.004 & 0     & 0     & -     \\ \hline
fin      & 0.073 & 0.114 & 0.089 & 0.044 & 0.203 & 0.072 & 0.045 & 0.17  & 0.071 & 0.049 & 0.177 & 0.076 & 0.033 & 0.196 & 0.056 & 0.073 & 0.12  & 0.091 & 0.037 & 0.047 & 0.041 & 0.007 & 0.144 & 0.014 \\ \hline
health   & 0.109 & 0.11  & 0.11  & 0.067 & 0.272 & 0.107 & 0.065 & 0.231 & 0.101 & 0.075 & 0.202 & 0.11  & 0.084 & 0.283 & 0.13  & 0.037 & 0.066 & 0.047 & 0.042 & 0.08  & 0.055 & 0.129 & 0.145 & 0.136 \\ \hline

\hline
         & \multicolumn{3}{|c|}{\cellcolor[HTML]{95A5A6}\textbf{TBATS}} & \multicolumn{3}{|c|}{\cellcolor[HTML]{95A5A6}\textbf{STL-ARMA-KF}} & \multicolumn{3}{|c|}{\cellcolor[HTML]{95A5A6}\textbf{STL}} & \multicolumn{3}{|c|}{\cellcolor[HTML]{95A5A6}\textbf{RobustSTL}} & \multicolumn{3}{|c|}{\cellcolor[HTML]{95A5A6}\textbf{STL-SDAR}} & \multicolumn{3}{|c|}{\cellcolor[HTML]{95A5A6}\textbf{STL-RobustKF}} & \multicolumn{3}{|c|}{\cellcolor[HTML]{95A5A6}\textbf{RobustPCA}} & \multicolumn{3}{|c|}{\cellcolor[HTML]{95A5A6}\textbf{PCA}} \\ \hline
\hline

         & {\textbf{Pr}} & {\textbf{Re}} & {\textbf{$F_1$}} & {\textbf{Pr}} & {\textbf{Re}} & {\textbf{$F_1$}} & {\textbf{Pr}} & {\textbf{Re}} & {\textbf{$F_1$}} & {\textbf{Pr}} & {\textbf{Re}} & {\textbf{$F_1$}} & {\textbf{Pr}} & {\textbf{Re}} & {\textbf{$F_1$}} & {\textbf{Pr}} & {\textbf{Re}} & {\textbf{$F_1$}} & {\textbf{Pr}} & {\textbf{Re}} & {\textbf{$F_1$}} & {\textbf{Pr}} & {\textbf{Re}} & {\textbf{$F_1$}} \\ \hline

yahoo-a1 & 0.106 & 0.295 & 0.156 & 0.225 & 0.378 & 0.282 & 0.133 & 0.454 & 0.206 & 0.111 & 0.525 & 0.183 & 0.372 & 0.23  & 0.284 & 0.299 & 0.313 & 0.306 & 0.038 & 0.585 & 0.071 & 0.062 & 0.191 & 0.094 \\ \hline
yahoo-a2 & 0.253 & 0.83  & 0.388 & 0.379 & 0.697 & 0.491 & 0.133 & 0.454 & 0.206 & 0.111 & 0.525 & 0.183 & 0.346 & 0.811 & 0.486 & 0.234 & 0.809 & 0.363 & 0.033 & 1     & 0.064 & 0.024 & 0.469 & 0.046 \\ \hline
yahoo-a3 & 0.678 & 0.995 & 0.806 & 0.865 & 0.999 & \cellcolor{blue!25}{\bf 0.927} & 0.242 & 1     & 0.39  & 0.307 & 0.999 & 0.47  & 0.537 & 0.999 & 0.698 & 0.468 & 0.986 & 0.635 & 0.023 & 0.996 & 0.045 & 0.184 & 0.686 & 0.291 \\ \hline
nab-ctr  & 0.053 & 0.857 & 0.1   & 0.048 & 0.786 & 0.09  & 0.02  & 0.929 & 0.038 & 0.014 & 1     & 0.027 & 0.079 & 0.714 & 0.142 & 0.07  & 0.786 & 0.128 & 0.009 & 1     & 0.018 & 0.002 & 0.071 & 0.003 \\ \hline
nab-twt  & 0.007 & 0.8   & 0.013 & 0.009 & 0.8   & 0.017 & 0.003 & 0.943 & 0.006 & 0.002 & 0.857 & 0.004 & 0.015 & 0.686 & 0.03  & 0.015 & 0.714 & 0.028 & 0.002 & 0.935 & 0.004 & 0.001 & 0.226 & 0.001 \\ \hline
nab-iot  & 0     & 0     & -     & 0.003 & 0.167 & 0.005 & 0     & 0.167 & 0.001 & 0.001 & 0.333 & 0.002 & 0     & 0     & -     & 0.001 & 0.167 & 0.003 & 0.001 & 1     & 0.002 & 0     & 0     & -     \\ \hline
fin      & 0.008 & 0.151 & 0.015 & 0.109 & 0.231 & 0.148 & 0.034 & 0.191 & 0.058 & 0.032 & 0.182 & 0.054 & 0.443 & 0.383 & 0.411 & 0.221 & 0.319 & 0.261 & -     & -     & -     & -     & -     & -     \\ \hline
health   & 0.109 & 0.096 & 0.102 & 0.103 & 0.11  & 0.107 & -     & -     & -     & -     & -     & -     & 0.104 & 0.105 & 0.105 & 0.057 & 0.05  & 0.053 & -     & -     & -     & -     & -     & -     \\ \hline

\hline
         & \multicolumn{3}{|c|}{\cellcolor[HTML]{95A5A6}\textbf{HOTSAX}} & \multicolumn{3}{|c|}{\cellcolor[HTML]{95A5A6}\textbf{RRA}} & \multicolumn{3}{|c|}{\cellcolor[HTML]{95A5A6}\textbf{DenStream}} & \multicolumn{3}{|c|}{\cellcolor[HTML]{95A5A6}\textbf{ClusTree}} & \multicolumn{3}{|c|}{\cellcolor[HTML]{95A5A6}\textbf{DBStream}} & \multicolumn{3}{|c|}{\cellcolor[HTML]{95A5A6}\textbf{IForest}} & \multicolumn{3}{|c|}{\cellcolor[HTML]{95A5A6}\textbf{MB$k$-means}} & \multicolumn{3}{|c|}{\cellcolor[HTML]{95A5A6}\textbf{OneSVM}} \\ \hline
\hline
         & {\textbf{Pr}} & {\textbf{Re}} & {\textbf{$F_1$}} & {\textbf{Pr}} & {\textbf{Re}} & {\textbf{$F_1$}} & {\textbf{Pr}} & {\textbf{Re}} & {\textbf{$F_1$}} & {\textbf{Pr}} & {\textbf{Re}} & {\textbf{$F_1$}} & {\textbf{Pr}} & {\textbf{Re}} & {\textbf{$F_1$}} & {\textbf{Pr}} & {\textbf{Re}} & {\textbf{$F_1$}} & {\textbf{Pr}} & {\textbf{Re}} & {\textbf{$F_1$}} & {\textbf{Pr}} & {\textbf{Re}} & {\textbf{$F_1$}} \\ \hline

yahoo-a1 & 0.783 & 0.386 & 0.517 & 0.675 & 0.488 & 0.566 & 0.569 & 0.702 & 0.628 & 0.594 & 0.686 & 0.637 & 0.646 & 0.537 & 0.586 & 0.46  & 0.531 & 0.493 & 0.563 & 0.68  & 0.616 & 0.58  & 0.724 & \cellcolor{blue!25}{\bf 0.644} \\ \hline
yahoo-a2 & 0.471 & 0.417 & 0.443 & 0.312 & 0.532 & 0.393 & 0.741 & 1     & \cellcolor{blue!25}{\bf 0.851}& 0.719 & 1     & 0.837 & 0.759 & 0.959 & 0.847 & 0.226 & 0.989 & 0.368 & 0.231 & 1     & 0.375 & 0.267 & 1     & 0.422 \\ \hline
yahoo-a3 & 0.431 & 0.328 & 0.372 & 0.105 & 0.244 & 0.147 & 0.411 & 0.514 & 0.457 & 0.466 & 0.755 & 0.577 & 0.784 & 0.492 & 0.605 & 0.486 & 0.672 & 0.564 & 0.471 & 0.433 & 0.451 & 0.603 & 0.737 & 0.663 \\ \hline
nab-ctr  & 0.148 & 0.857 & 0.253 & 0.141 & 1     & 0.248 & 0.181 & 0.929 & 0.302 & 0.197 & 0.929 & 0.325 & 0.224 & 0.929 & 0.361 & 0.3   & 0.857 & \cellcolor{blue!25}{\bf 0.444} & 0.169 & 0.786 & 0.278 & 0.162 & 0.786 & 0.268 \\ \hline
nab-twt  & 0.169 & 0.571 & \cellcolor{blue!25}{\bf 0.261} & 0.16  & 0.6   & 0.253 & 0.038 & 0.971 & 0.074 & 0.037 & 0.971 & 0.071 & 0.066 & 0.914 & 0.123 & 0.025 & 0.971 & 0.049 & 0.034 & 0.943 & 0.066 & 0.023 & 0.714 & 0.045 \\ \hline
nab-iot  & 0.107 & 0.5   & \cellcolor{blue!25}{\bf 0.176} & 0.054 & 0.5   & 0.097 & 0.01  & 0.5   & 0.02  & 0.012 & 0.5   & 0.023 & 0.012 & 0.333 & 0.023 & 0.009 & 0.667 & 0.018 & 0.009 & 0.667 & 0.018 & 0.009 & 0.667 & 0.018 \\ \hline
fin      & 0.932 & 0.449 & 0.606 & 0.952 & 0.663 & 0.781 & 0.806 & 0.908 & 0.854 & 0.816 & 0.91  & \cellcolor{blue!25}{\bf 0.861} & 0.888 & 0.759 & 0.819 & 0.808 & 0.78  & 0.794 & 0.686 & 0.936 & 0.792 & 0.749 & 0.91  & 0.822 \\ \hline
health   & 0.949 & 0.335 & 0.495 & 0.788 & 0.387 & 0.519 & 0.775 & 0.398 & 0.526 & 0.827 & 0.512 & \cellcolor{blue!25}{\bf 0.633} & 0.9   & 0.47  & 0.618 & 0.858 & 0.3   & 0.445 & 0.79  & 0.349 & 0.484 & 0.679 & 0.273 & 0.389 \\

\hline
\hline
\end{tabular}
}
\vspace{1mm}
\caption{Precision, Recall and $F_1$-scores for all the techniques. For each dataset, the most accurate technique is highlighted.}
\vspace{-4mm}
\label{tab:accuracytable}
\end{table*}

\begin{dinglist}{112}

\item {\bf Robustness and False Positives:} Techniques such as {\em med-mad}
      surface a higher number of anomalies, which improves recall at the
      expense of precision as evident from \tabref{tab:accuracytable}.
      %
      For best accuracy, we recommend to use robust techniques with CFAR such
      as {\em GESD}. From \tabref{tab:accuracytable} we note that {\em GESD}
      outperforms most other statistical techniques. On the other hand, from
      \figref{fig:accspeed} we note that {\em GESD} has the highest runtime
      amongst the statistical techniques.
      \vspace{0.8mm}
\item {\bf Model Building:} Estimating model parameters in the presence of
      anomalies can potentially impact accuracy adversely if the technique is
      not robust. This is observed from \figref{fig:robustness} wherein an
      extreme anomaly biases the model obtained via {\em SARMA} thereby inducing
      false positives in the {\em nab-ctr} dataset.
      {\em STL} is susceptible to this as well. In contrast, {\em Robust STL}
      effectively down-weights the anomalies during model parameter estimation.
      From \figref{fig:robustness} we also note that pattern mining techniques
      such as {\em DenStream} are more robust to anomalies, as they do not fit
      a parametric model.
      %
      %
      \vspace{0.8mm}
\begin{figure}[!h]
\centering
\includegraphics[width=\linewidth, height=2in]{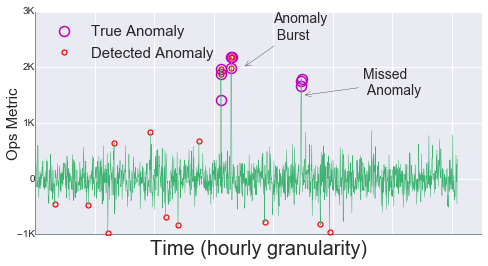}
\vspace*{-6mm}
\caption{Peformance of {\em t-digest} anomaly bursts}
\vspace*{-4mm}
\label{fig:tdigest_hysterisis}
\end{figure}

\item {\bf Anomaly Bursts:} It is not uncommon to observe bursts of anomalies
      in production data. In such a scenario, the accuracy of a technique is
      driven by how soon the technique adapts adapts to the new ``normal". If
      a burst is long enough, then most techniques do adapt but with different
      lag.
      CFAR techniques such as {\em t-digest} and {\em GESD} fair quite poorly
      against anomaly bursts. For instance, in the {\em health} data set, the
      anomalies happen in bursts and a CFAR system puts an upper bound on the
      number of anomalies, thereby missing many of them.
      Having said that, this can also be advantageous as exemplified by the
      {\em nab-ctr} data set wherein there are a few spaced out 
      anomalies.
      \tabref{tab:accuracytable} shows that {\em t-digest} improves both precision
      and recall. \figref{fig:tdigest_hysterisis} illustrates an operations time
      series that highlights why {\em t-digest} does not surface anomaly bursts.
      %
      \vspace{0.8mm}
\begin{figure}[!h]
\centering
\includegraphics[width=\linewidth, height=2.5in]{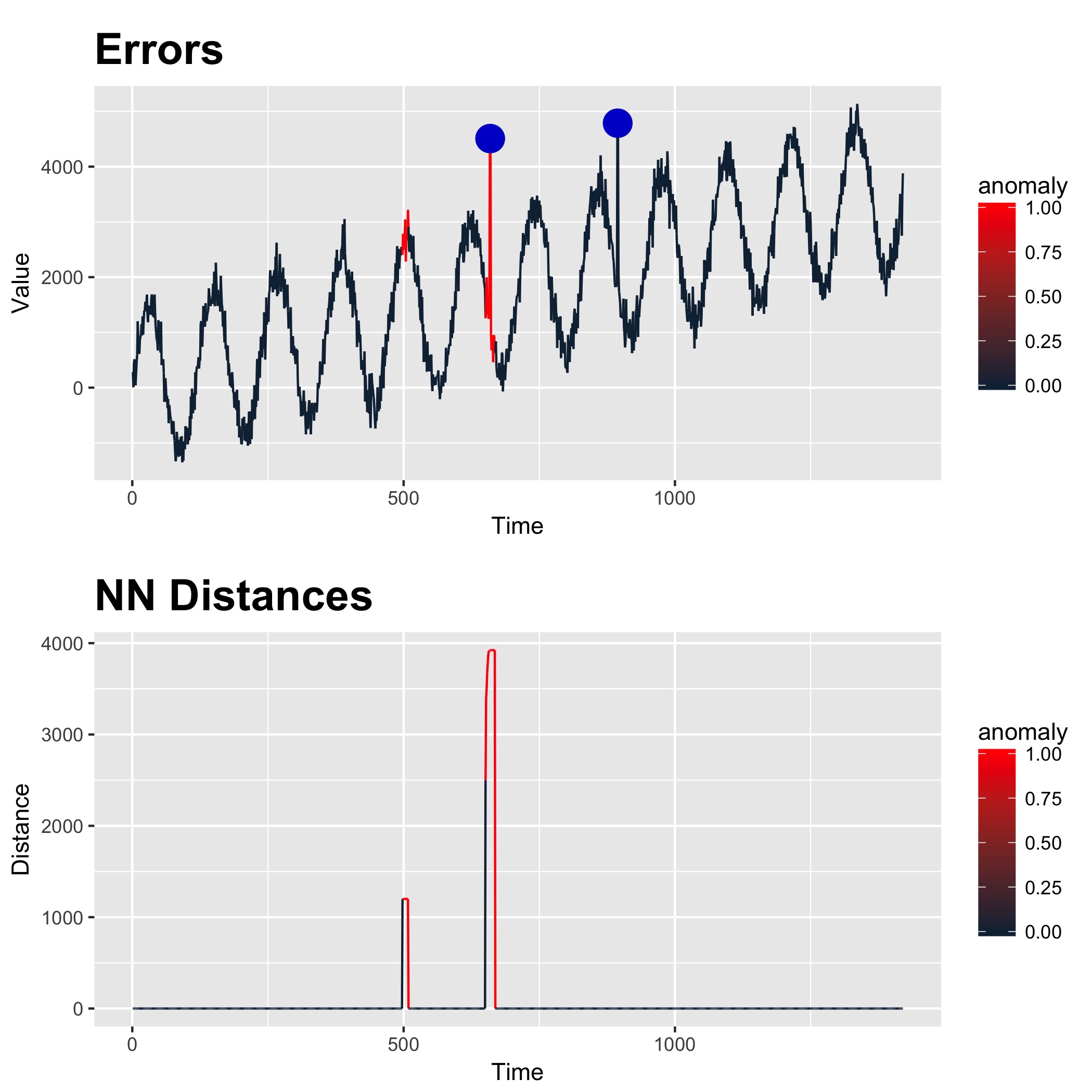}
\vspace*{-6mm}
\caption{HOTSAX similar anomalies}
\vspace*{-4mm}
\label{fig:hotsax_anomaly}
\end{figure}

\item {\bf Unique Pattern Anomalies:} The performance of {\em HOTSAX} and {\em
      RRA} is abysmal on the {\em yahoo-a2} and {\em yahoo-a3} data sets. This
      is because these synthetic data sets comprise of many similar anomalies.
      Both {\em HOTSAX} and {\em RRA} are not robust to the presence of such
      similar anomalies as the anomaly score is based on the nearest neighbor
      distance. 
      \figref{fig:hotsax_anomaly} highlights how similar anomalies may be
      missed. In contrast, {\em DenStream} and {\em DBStream} are able to detect
      self-similar anomalies as they create micro-clusters of similar anomalies.
      \vspace{-0.2mm}
\begin{figure}[!h]
\centering
\includegraphics[width=\linewidth, height=2.5in]{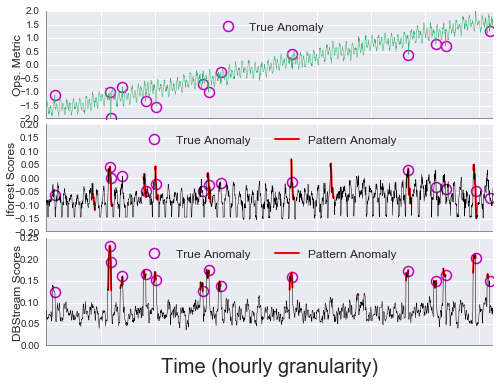}
\vspace{-7mm}
\caption{Anomaly Score Separation}
\vspace*{-4mm}
\label{fig:anom_score}
\end{figure}

\item {\bf Scale of distance measure:} Accuracy of an anomaly detection technique
      is a function of the distance measure used to differentiate normal vs.
      anomalous data points.
      For instance, let us consider {\em IForest} and {\em DBStream} (refer to
      \figref{fig:anom_score}). The latter creates much better separation between
      normal and anomalous data points. This can be ascribed to anomaly score in
      {\em IForest} being the depth of the leaf in which the point resides, which
      is analogous to the {\em log} of the distance.
      %

\end{dinglist}

\vspace{-1mm}
\subsection{Model Selection}

\noindent
As mentioned earlier, the analysis presented in this paper should serve
as a guide for selection of the ``best" anomaly detection technique. In
general, model selection is a function of the application domain and
latency requirement.
\tabref{tab:recs} enlists the various application domains, the attributes
exhibited by the datasets in these domains and the ``best" algorithms for
a given latency requirement (according to the accuracy-speed trade-offs
discussed in the previous section).

\begin{table}[ht]
\centering
\large
\newcolumntype{L}[1]{>{\raggedright\let\newline\\\arraybackslash\hspace{0pt}}m{#1}}
\newcolumntype{C}[1]{>{\centering\let\newline\\\arraybackslash\hspace{0pt}}m{#1}}
\newcolumntype{R}[1]{>{\raggedleft\let\newline\\\arraybackslash\hspace{0pt}}m{#1}}
\resizebox{\linewidth}{!}{
\begin{tabular}{C{2.5cm}|C{2cm}|C{3cm}|C{2.5cm}|C{2.5cm}|C{2.5cm}}
\hline
\hline
  {\cellcolor[HTML]{95A5A6}\textbf{Application Domain}} & {\cellcolor[HTML]{95A5A6}\textbf{DataSets}} & {\cellcolor[HTML]{95A5A6}\textbf{Attributes}} & {\cellcolor[HTML]{95A5A6} \textbf{\textless $1$ msec}} & {\cellcolor[HTML]{95A5A6} \textbf{$1$-$10$ msec}} & {\cellcolor[HTML]{95A5A6} \textbf{\textgreater $10$ msec}} \\ \hline
  Hourly Operations & yahoo-a1, yahoo-a2, yahoo-a3 & LS, VC, SLC, Noisy & STL-SDAR/ STL-RobustKF & DenStream & DenStream \\ \hline
  Advertising & nab-ctr & No Anomaly Bursts & t-digest & DBStream & IForest \\ \hline
  Hourly IoT & nab-twt, nab-iot & Unique Anomalies & med-mad & DBStream & HOTSAX \\ \hline
  Financial & fin & LS, VC & SDAR & DenStream & ClusTree \\ \hline
  Healthcare & health & SJ, LS & -- & DBStream & DBStream/ HOTSAX \\ \hline
  Minutely Operations & ops & LS, VC, SLS, SLD, Large SP & med-mad & DBStream & DBStream \\
\hline
\hline
\end{tabular}
}
\vspace{1mm}
\caption{Best Techniques For Different Latency Requirements.
{\bf SP}: Seasonal Period,
{\bf SJ}: Seasonal period Jitter,
{\bf LS}: Level Shift,
{\bf VC}: Variance Change,
{\bf SLD}: Seasonal Level Drift,
{\bf SLS}: Seasonal Level Shift}
\vspace{-9mm}
\label{tab:recs}
\end{table}

For applications with latency requirement <$1$ msec, the use of pattern and
machine learning based anomaly detection techniques is impractical owing to
their high computation requirements. Although techniques geared towards
detecting point anomalies can be employed, the ``best" technique is highly
dependent on the attributes.
For instance, {\em STL-SDAR} accurately detects anomalies for operations time
series that exhibit non-stationarities such as $LS$, $VC$. On the other hand,
in the case of minutely operations time series which typically tend to exhibit
long Seasonal Periods (SP), {\em STL} becomes expensive, and hence a simpler
technique such as {\em med-mad} can potentially be used to meet the latency
requirement.
One can use {\em SDAR} for financial time series even if the latency requirements
are strict. This stems from the fact that these series (mostly) do not exhibit
seasonality, and hence {\em STL} is not a bottleneck.

\begin{figure}[!h]
\centering
\includegraphics[width=\linewidth, height=2.5in]{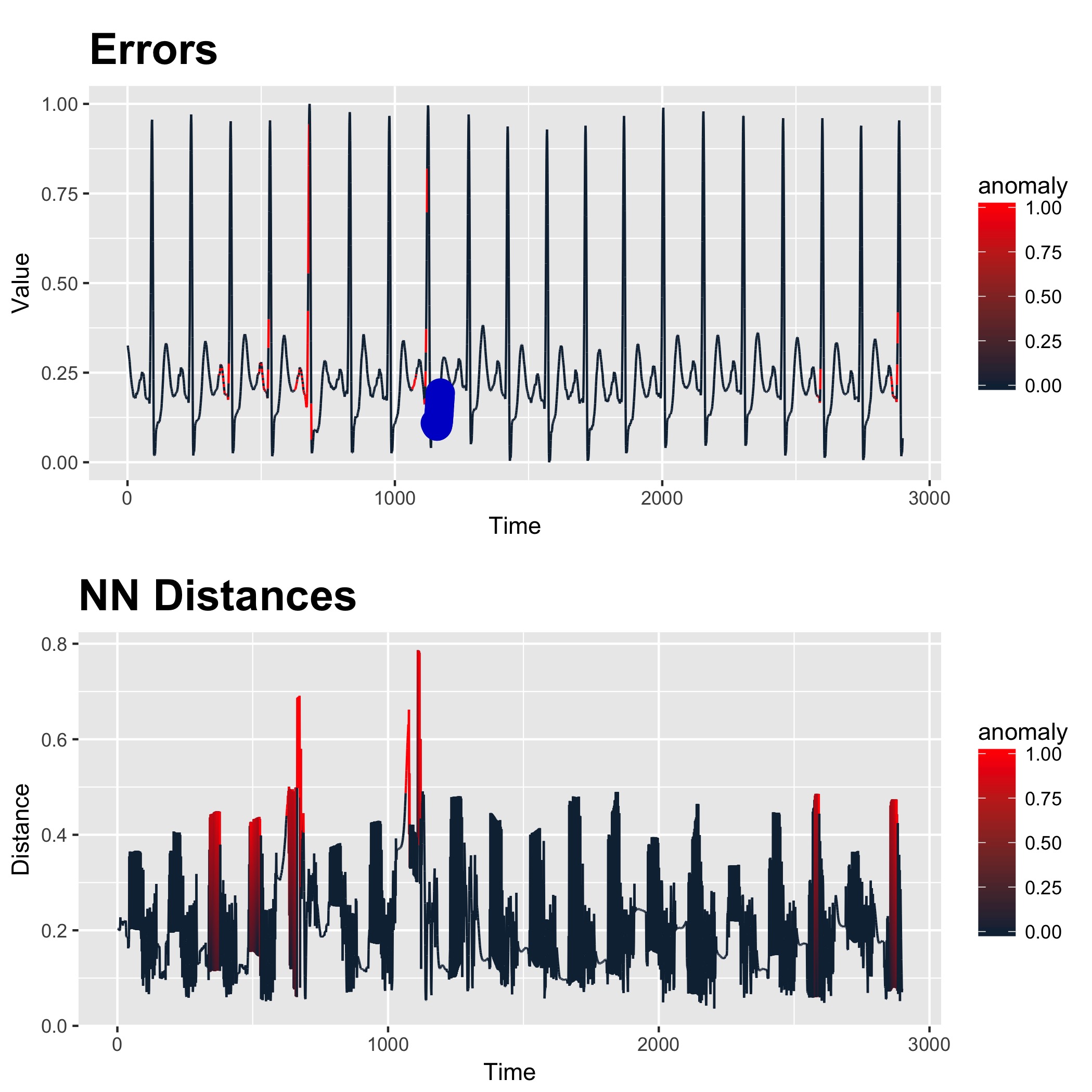}
\vspace*{-6mm}
\caption{HOTSAX and DBStream for {\em health-qtdb/0606}}
\vspace*{-7mm}
\label{fig:hotsax_dbstream}
\end{figure}

When latency requirement is in the range of $1$-$10$ msec, micro-clustering
techniques like {\em DBStream} and {\em DenStream} outperform all others.
Although {\em DBStream} marginally outperforms {\em DenStream}, in the presence
of noisy time series {\em DenStream} is more effective. It has been shown in
prior work \cite{pattern_hotsax_Keogh_2005, pattern_rra_2015} that techniques
such as {\em HOTSAX} are effective in finding anomalies in ECG data in an offline
setting. From \figref{fig:accspeed} we note that the detection runtime is over
$100$ msec, which is significantly larger than the {\em TG} for ECG series.
{\em DBStream} on the other hand is much faster ($5$-$10$msec) and can detect
the same set of anomalies as {\em HOTSAX}, refer to \figref{fig:hotsax_dbstream}.
Having said that, {\em DBStream} does surface more false positives than {\em
HOTSAX}. A post processing step can help reduce the false positive rate. It
should be noted that none of the aforementioned techniques satisfy the <$1$
msec latency requirement for the health time series.

Finally, when the latency requirements are >$10$msec, a wide range of anomaly
techniques can be leveraged. In many cases, {\em DBStream} is still the most
accurate technique. When an application has very few unique anomalies, {\em
HOTSAX} usually is the most effective, as is the case with the datasets {\em
nab-twt} and {\em nab-iot}.

\vspace*{-1mm}



\section{Conclusion} \label{sec:conc}
\vspace*{-1mm}

\noindent
In this paper, we first presented a classification of {\em over 20} anomaly
detection techniques across seven dimensions (refer to \tabref{tab:taxonomy}).
Next, as a first, using {\em over 25} real-world data sets and real hardware,
we presented a detailed evaluation of these techniques with respect to their
real-timeliness and performance -- as measured by  {\em precision, recall}
and $F_1$ score.
We also presented a map of their accuracy-runtime trade-off.  Our experiments
demonstrate that the state-of-the-art anomaly detection techniques are applicable
for data streams with $1$ msec or higher granularity, highlighting the need for 
faster algorithms to support the use cases mentioned earlier in \secref{sec:intro}.
Last but not least, given an application domain and latency requirement, based 
on empirical evaluation, we made recommendations for the ``best" technique for
anomaly detection.

As future work, we plan to extend our evaluation to other data sets such as
live data streams as exemplified by Facebook Live, Twitter Periscope video
applications and other live data streams on platforms such as Satori.
\vspace{2mm}

\vspace*{-12mm}
{
\vspace*{+10mm}
\fontsize{6}{0.16cm}%
\selectfont%
\bibliographystyle{ACM-Reference-Format}

}

\end{document}